%% file: main2.tex
\documentclass[12pt]{iopart}
\usepackage[utf8]{inputenc} 
\usepackage[T1]{fontenc}    
\usepackage{hyperref}       
\usepackage{url}            
\usepackage{booktabs}       
\usepackage{nicefrac}       
\usepackage{microtype}      
\usepackage{float}
\usepackage{amsmath}
\usepackage{amssymb}
\usepackage[caption=false]{subfig}
\usepackage{multirow}
\usepackage{graphicx}
\usepackage{placeins}
\usepackage{xcolor}
\usepackage{appendix}
\usepackage{algorithm}
\usepackage[noend]{algpseudocode}
\usepackage{lineno}

\newcommand{\partialD}[2]{\frac{\partial #1}{\partial #2}}
\newcommand{\params}{\mathbf{\theta}}
\newcommand{\argmin}{\mathop{\mathrm{argmin}}\limits}

\newcommand\sbullet[1][.5]{\mathbin{\vcenter{\hbox{\scalebox{#1}{$\bullet$}}}}}

\usepackage{iopams}  
\begin{document}


\pagebreak
\title[Robust Learning of Physics Informed Neural Networks]{Recipes for when Physics Fails: Recovering Robust Learning of Physics Informed Neural Networks}

\author{Chandrajit Bajaj$^1$, Luke McLennan$^1$, Timothy Andeen$^2$, and Avik Roy$^3$\footnote{corresponding author}}

\address{$^1$ Department of Computer Science \& Oden Institute for Computational Engineering and Sciences,
The University of Texas at Austin, Austin, TX 78712}
\address{$^2$ Department of Physics,
The University of Texas at Austin, Austin, TX 78712}
\address{$^3$ Center for AI Innovation, National Center for Supercomputing Applications,
University of Illinois at Urbana Champaign, Urbana, IL 61801}
\eads{avroy@illinois.edu}

\vspace{10pt}
\begin{indented}
\item[]November 2022
\end{indented}

\begin{abstract}
Physics-informed Neural Networks (PINNs) have been shown to be effective in solving partial differential equations by capturing the  physics induced constraints as a part of the training loss function. This paper shows that a PINN can be sensitive to errors in training data and overfit itself in dynamically propagating these errors over the domain of the solution of the PDE. It also shows how physical regularizations based on continuity criteria and conservation laws fail to address this issue and rather introduce problems of their own causing the deep network to converge to a physics-obeying local minimum instead of the global minimum. We introduce Gaussian Process (GP) based smoothing that recovers the performance of a PINN and promises a robust architecture against noise/errors in measurements. Additionally, we illustrate an inexpensive method of quantifying the evolution of uncertainty based on the variance estimation of GPs on boundary data. 
  Robust PINN performance is also shown to be achievable by choice of sparse sets of inducing points based on sparsely induced GPs. We demonstrate the performance of our proposed methods and compare the results from existing benchmark models in literature for time-dependent Schr\"odinger and Burgers' equations.
\end{abstract}

%
\vspace{2pc}
\maketitle
%
%

\input{tex-paper/1Intro}
\input{tex-paper/2PINN}
\input{tex-paper/3ErrorProp}
\input{tex-paper/4GPSmoothing}

\input{tex-paper/5AddnlEx}
\input{tex-paper/6Conclusion}

\FloatBarrier

\section*{References}
\bibliographystyle{unsrt}
\bibliography{main2}

\end{document}

%% file: tex-paper/1Intro.tex
\section{Introduction}\label{sec:intro}

Neural Networks (NNs) are finding ubiquitous applications in fundamental sciences. Their abilities to perform classification and regression over large and complicated datasets are making them extremely useful for a variety of purposes, including modeling  molecular dynamics modeling, nonlinear dynamical system design and control, and other many body interactions~\cite{MolecularML,NonlinearSystemNN,ManyBodyNN}. In many cases, these important problems in physics and engineering are posed in terms of static and time dependent partial differential equations (PDEs).  Analytical solutions to PDEs are scarce. Moreover, PDEs are difficult to solve and require computationally intensive, and highly pre-conditioned numerical linear solvers~\cite{benzi2002preconditioning}. Popular discretization methods, Finite Difference Method (FDM), and even Finite Element Method (FEM) are used to obtain point wise or piece-wise linear estimates over a fine grid or meshed domains of interest~\cite{grossmann2007numerical}. Although NN-based approximations to differential equations have been explored for some time~\cite{dissanayake1994neural,aarts2001neural,hayati2007feedforward}, interest in such approaches have been reinvigorated recently due to significant improvement in computational platforms that support fast forward and backward gradient propagation utilizing automatic differentiation~\cite{baydin2018automatic}. Rapid progress has been seen in NN-assisted solutions of Ordinary Differential Equations (ODEs)~\cite{ANNforSolvingDE}. Novel architectures like NeuralODEs~\cite{NeuralODE} have been proposed to harness the power of blackbox ODE solvers in conjunction with continuous-depth residual Neural Networks (Resnets) using the method of adjoints~\cite{adjoint}, while models like ODE2VAEs use variational auto-encoder architectures to learn functions and derivatives via latent space embeddings similar to Cauchy boundary conditions~\cite{ODE2VAE}. 

Solving PDEs using NNs has also seen significant attention by the development of  Physics Informed Neural Networks (PINNs)~\cite{pinn-main}. Trained PINNs have been shown to be effective in solving time dependent partial differential equations for a given set of Cauchy boundary conditions over a finite spatio-temporal domain. They exploit a deep neural network's ability as universal function approximators~\cite{approximator, approximator-2}. Many variants of the trainable PINN architecture have been explored to exploit structure, quality and speed of convergence, and dimensional scalability~\cite{pinn-fractional,pinn-bayes,pinn-adaptive,pinn-adversarial,pinn-parareal,cpinn-2,Shin_2020}. However, one of the least explored areas is the robustness of trainable PINNs for various noisy data scenarios and conditions. While most PINN architectures in literature assume perfectly known boundary data, in many practical applications, this data comes from regulated and calibrated measurement processes and is subject to uncertainties or errors pertaining to the limitations of the measurement system or the stochastic nature of the physical processes themselves. Most PINN architectures utilize a finite and small collection of training data on the domain boundary.  Given the typical small size of this training data and NNs ability to capture arbitrary non-linearity, vanilla PINNs can often propagate these errors or uncertainties in an unstable fashion. Such unregulated error propagation can significantly limit the applicability of PINNs as industrial strength numerical approximators of PDEs.

In this paper, we extensively investigate the problem of error propagation in PINNs. In Section~\ref{sec:pinn} we introspect the architecture of a PINN, and how it responds when data on the initial timeslice is corrupted with noise. The notion of PINN robustness is thus tied to the PINNs ability to preserve solution features under such noisy perturbations.  We analyze examples of  learning time-varying non-linear Schr\"odinger, and the non-convective fluid flow  Burgers' equations. We further show the impact of introducing regularization based on continuity criteria inspired from conservative PINN (cPINN) architectures~\cite{pinn-adaptive,cpinn-2} through attempts to satisfy conservation laws.
In Section~\ref{sec:gp-pinn} we provide details of a Gaussian Process smoothed PINN (GP-smoothed PINN), and its sparse variant and compare its performance against most typical and popular PINN architectures. We demonstrate how these twin GP-smoothed PINNs recover the intended solution and can outperform methods realizing continuity or conservation regularizers ~\cite{pinn-conservative} as well as better possess the ability to control uncertainty propagation compared to uncertainty quantification methods proposed for example in~\cite{pinn-adversarial}. We briefly examine the efficiency role of sparse inducing points and also demonstrate  the importance and choice  of various kernels for achieving best model selection in Gaussian process training to prevent data driven under- or overfitting.

%% file: tex-paper/2PINN.tex
\section{Robustness of PINNs}\label{sec:pinn}
In this section, we review the PINN and cPINN architectures and explain with different examples how such models fail to capture the essence of robustness. We also investigate multiple physics-inspired regularization schemes and identify their limitations in addressing the issue of robustness for usual PINN architectures.

\subsection{Review of PINN Architectures}

A partial differential equation that determines spatio-temporal evolution of a set of scalar (real or complex) fields, collectively represented by $u(\vec{x})$, can be expressed as 
\begin{equation}
\mathcal{N}\left[u(\vec{x}), f(\vec{x})\right] = 0
\label{eqn:PDE-general}
\end{equation}
where $\vec{x}$ defines a $n$ dimensional spatial or saptio-temporal coordinate system, defined on the domain $\vec{x} \in \mathcal{D} \subset \mathbb{R}^n$. $\mathcal{N}$ represnts a set of known, finite-order, differential operators  and $f(\vec{x})$ is the source function, usually known as analytical expression in problems intended to solve a forward PDE problem. Eq. \ref{eqn:PDE-general} is subject to a set of boundary conditions,
\begin{equation}
\mathcal{B}\left[u(\vec{x} \in \partial D)\right] = 0
\label{eqn:BC-general}
\end{equation}
The general idea of a PINN~\cite{pinn-main} is to obtain an approximation of the field $u(\vec{x}) \approx \tilde{u}(\vec{x})$ by a deep neural network that can solve the system of equations \ref{eqn:PDE-general}, subject to \ref{eqn:BC-general}. 
\begin{equation}
\tilde{u}(\vec{x}) = \mathbf{NN}_\params\left( \vec{x}; \mathcal{U}_B, \mathcal{U}_C,  \mathcal{U}_D \right)
\label{eqn:PINN-def}
\end{equation}
where, 
\begin{itemize}
\item
$\mathcal{U}_B = \{ (\vec{x}_i^b , \mathcal{B}[u(\vec{x}_i^b)] )_{i=1}^{N_b} \}$ represents a set of samples on the domain boundary $\partial\mathcal{D}$, 
\item
$\mathcal{U}_C = \{ (\vec{x}_i^c , f(\vec{x}_i^c) )_{i=1}^{N_c} \}$ is a set of measurements that enforces the PDE physics of Eq.~\ref{eqn:PDE-general} on the neural net (Eq.~\ref{eqn:PINN-def}), and
\item
$\mathcal{U}_D = \{ (\vec{x}_i^d , u(\vec{x}_i^d) )_{i=1}^{N_d} \}$ is a set of direct measurements. Although, $\mathcal{U}_D$ is not necessary for training a plain PINN as in Ref.~\cite{pinn-main}, they are often used for training networks for targeted simulation.
\end{itemize}

The parameters of the deep network $(\params)$ in Eq.~\ref{PINN} are obtained by minimization of the loss function,
\begin{equation}
\params^* = \argmin_\params\mathcal{L}_{PINN}
\end{equation}
where the loss function can be decomposed as,
\begin{equation}
\mathcal{L}_{PINN} = \alpha_{BC}\mathcal{L}_{BC} + \alpha_{PDE}\mathcal{L}_{PDE} + \alpha_{D}\mathcal{L}_{D}
\label{eqn:loss-pinn}
\end{equation}
Here, $\mathcal{L}_{BC}$ is the MSE loss calculated over $\mathcal{U}_B$, enforcing the NN to approximate the boundary condition,
\begin{equation}
\mathcal{L}_{BC} = 
\frac{1}{N_b}\sum_i\left|\mathcal{B}[\tilde{u}(\vec{x}_i^b)] \right|^2
\label{eqn:loss-bc}
\end{equation}
$\mathcal{L}_{PDE}$ is the MSE loss calculated over $\mathcal{U}_C$, enforcing the physics on the NN,
\begin{equation}
\mathcal{L}_{PDE} = \frac{1}{N_c}\sum_i\left|\mathcal{N}[\tilde{u}(\vec{x}_i^c), f(\vec{x}_i^c) ] \right|^2
\label{eqn:loss-pde}
\end{equation}
and finally, $\mathcal{L}_{D}$, if employed, determines the loss with respect to observation,
 \begin{equation}
\mathcal{L}_{D} = \frac{1}{N_d}\sum_i\left|\tilde{u}(\vec{x}_i^d) - u(\vec{x}_i^d) \right|^2
\label{eqn:loss-data}
\end{equation}
The $\alpha_{()}$ parameters in Eq.~\ref{eqn:loss-pinn} are penalty parameters that determine the relative strength of the regularizing terms in the loss function. Authors in~\cite{pinn-main} assign $\alpha_{()} = 1$ identically, but alternate choices have been explored in other works~\cite{pinn-conservative, pinn-variational}.


\begin{figure}
    \centering
    \subfloat[]{
        \includegraphics[width=0.3\textwidth]{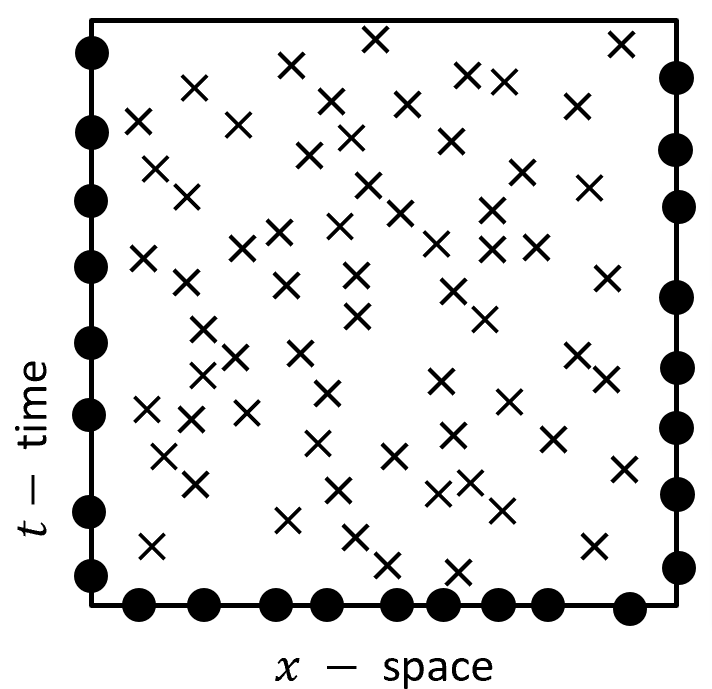}
        \label{fig:pinn-domain}
    }
    \subfloat[]{
        \includegraphics[width=0.3\textwidth]{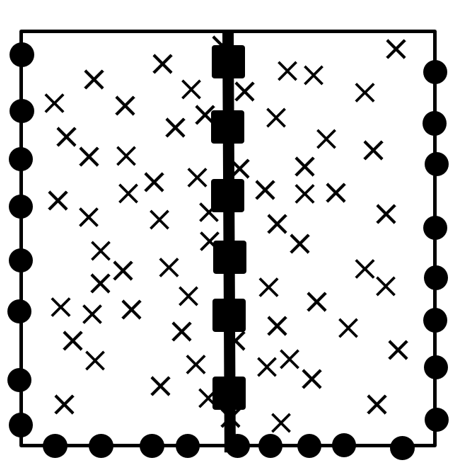}
        \label{fig:cpinn-domain}
    }
    \subfloat[]{
        \includegraphics[width=0.3\textwidth]{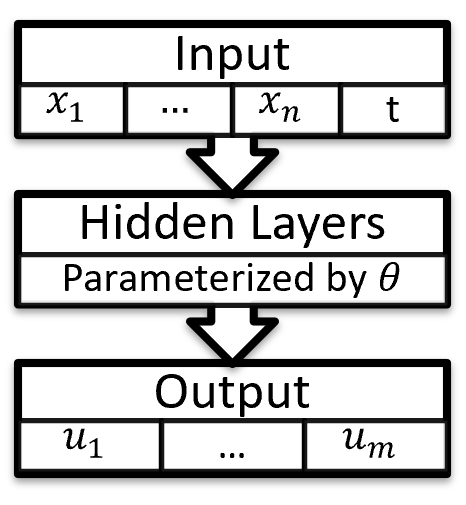}
        \label{fig:pinn-arch}
    }
    \caption{A visualization of the domains of \protect\subref{fig:pinn-domain} a PINN and \protect\subref{fig:cpinn-domain} a cPINN  with boundary points ($\sbullet[0.75]$), collocation points ($\times$), and interface points ($\blacksquare $) for a spatio-temporal domain with one spatial dimension. Figure \protect\subref{fig:pinn-arch} shows the model architecture diagram of a PINN for a generic $n+1$ dimensional spatio-temporal domain solving for a set of coupled fields $u_1, ... , u_m$.}
    \label{fig:pinn-domain-arch}
\end{figure}

\subsubsection{Conservative PINNs (cPINNs)}

Variations of PINN architectures have been explored in a number of recent works. For example, authors in~\cite{pinn-parareal} explore a multi-staged PDE solver for a long range solution and Ref.~\cite{pinn-adaptive} introduces adaptive, hyperparameterized activation functions that would accelerate the convergence of such networks.

Although the training of PINNs doesn't strictly require a discretized grid of evaluation points as often required by traditional numerical techniques like finite difference method and finite element methods, they can certainly benefit from such grid structures by parallelizing the training of PINNs over a collection of subdomains. Hence, the domain of integration and its boundary are divided into subdomains, i.e. 

\begin{equation}
\mathcal{D} = \bigcup_{i=1}^{K}d_i,\quad\quad
\mathcal{\partial D} = \bigcup_{i=1}^{K}\partial d_i, \quad\quad
\tilde{u} =  \bigcup_{i=1}^{K}\tilde{u}_i
\end{equation}

An additional benefit to such parallelized structure is to include flux continuity at subdomain boundaries in the loss function, providing additional safeguard against error propagation. In other words, the loss function in Eqn.~\ref{eqn:loss-pinn} is modified as
\begin{equation}
\mathcal{L}_{cPINN} = \sum_{j=1}^d \mathcal{L}_{PINN}^j + \alpha_{I}\mathcal{L}_{I}^j 
\label{eqn:loss-cpinn}
\end{equation}
where $\mathcal{L}_{PINN}^j$ is the PINN loss for the $j$-th subdomain as defined in Eq.~\ref{eqn:loss-pinn}. The other term, $\mathcal{L}_{I}^j$ acts as a regularizer that enforces functional and flux continuity at the interface of $j$-th subdomain interface.
\begin{equation}
\mathcal{L}_{I}^j = \frac{1}{N_{Ij}}\sum_{i=1}^{N_{Ij}}\left(
\left|\tilde{u}_j(\vec{x}_i^j) - \tilde{u}_{j+1}(\vec{x}_i^{j}) \right|^2 +
\left|\nabla \tilde{u}_j(\vec{x}_i^j)\cdot \mathbf{n_i^j}- \nabla \tilde{u}_{j+1}(\vec{x}_i^{j})\cdot \mathbf{n_i^{j+1}} \right|^2 \right)
\label{eqn:loss-I} 
\end{equation}
here, $\{\vec{x}_i^j\}$ is a collection of points on the interface of $j$-th and $j+1$-th subdomains and  $\mathbf{n_i^j}$ is the unit vector normal to the interface of  $j$-th subdomain at the location of $\vec{x}_i^j$.

Figure~\ref{fig:pinn-domain-arch} shows a schematic representation of the different spatio-temporal regions that contribute to evaluating the loss function in Eqn.~\ref{eqn:loss-pinn} along with a generic architecture for such models.

%% file: tex-paper/3ErrorProp.tex
\subsection{Error Propagation Through PINNs} \label{sec:pinn-err}

An NN can approximate non-linear functions with increasing degrees of accuracy. It is typically expected in the case of a PINN that $N_b \ll N_c$ i.e. the size of the training data on the domain boundary is usually much smaller than the size of the physics-enforcing collocation points. This unavoidable feature of PINNs make them susceptible to overfitting on the boundary and eventually propagate those errors across the domain. In the following subsections, we investigate the physical nature of error propagation through PINNs and the impact of different regularizations on training the PINN architecture. We will use two popular examples that have been widely used in the literature to illustrate this issue of error propagation.


\subsubsection{Nonlinear Schr\"{o}dinger equation}

We consider the example considered in Ref.~\cite{pinn-main} of a nonlinear Schr\"{o}dinger partial differential equation which describes the spatio-temporal evolution of a 1D complex field $h(x,t) = u(x,t) + iv(x,t)$ as

\begin{equation}
i\partialD{h}{t} + \frac{1}{2} \partialD{^2 h}{x^2} + |h|^2h = 0
\label{eqn:pde-schrod}
\end{equation}
which can also be interpreted as a set of coupled partial differential equations given as
\begin{align}
-\partialD{v}{t} + \frac{1}{2} \partialD{^2 u}{x^2} + (u^2 + v^2)u &= 0 \nonumber \\
\partialD{u}{t} + \frac{1}{2} \partialD{^2 v}{x^2} + (u^2 + v^2)v &= 0
\label{eqn:pde-schrod-2}
\end{align}
The domain boundary is defined as $(x,t) \in [-5, 5] \times [0, \frac{\pi}{2}]$. The boundary conditions can be classified as 
\begin{itemize}
\item[i.] Initial condition; the known value of the field on the initial time slice given as a collection of measurements $\{(x_j, h(x_j,0)_{j=0}^{N_{b,t}}\}$. Unbeknownst to the NN, these measurements are taken from the analytical solutions with possible sources of additive corruption- 
\begin{align}
u(x_j, 0) &= 2\mathrm{sech}(x_j) + \Theta_u\epsilon_i^u \nonumber \\
v(x_j, 0) &= \Theta_v\epsilon_i^v 
\label{eqn:schrod-IC}
\end{align}
where $\epsilon^u, \epsilon^v$ are randomly chosen from a normal distribution with $\mathcal{N}(0, \sigma^2)$. The parameters $\Theta_u$ and $\Theta_v$ represent acceptance of errors which are set to 0 (1) for error-free (error-inclusive) initial conditions.
\item[ii.] Periodic boundary condition on spatial slices enforced on a discretized spatial boundary. A total of $N_{b,s}$ points are chosen to enforce the following spatial boundary conditions  
\begin{align}
h(+5, t) &= h(-5, t) \nonumber \\
\partialD{h}{x} (+5,t) &= \partialD{h}{x} (-5,t)
\label{eqn:schrod-PBC}
\end{align}
\end{itemize}
The loss function is constructed according to Eqn. \ref{eqn:loss-pinn} with $\alpha_{()} = 1.0$. To set the benchmark for the performance of a PINN for this problem, we solved for the PDE with error-free boundary data ($\Theta_u = \Theta_v = 0$). We train a fully connected MLP with 6 hidden layers and 70 nodes per layer, with two inputs corresponding to the space and time coordinate and two outputs corresponding to the real and imaginary parts of the complex field. The MLP was trained for $N_b = 100$ points on the domain boundary, 50 of which were taken from uniformly sampling the space coordinate $(x)$ on the initial timeslice to impose the initial condition in Eq. \ref{eqn:schrod-IC}, and the 50 points were taken on from a uniform distribution on the time coordinate to impose the periodic boundary condition in Eq. \ref{eqn:schrod-PBC}. A fine grid of $N_c = 20000$ collocation points was chosen to impose the physics of the PDE.

\begin{figure}
\centering
  \subfloat[]{
  \includegraphics[width=0.24\textwidth]{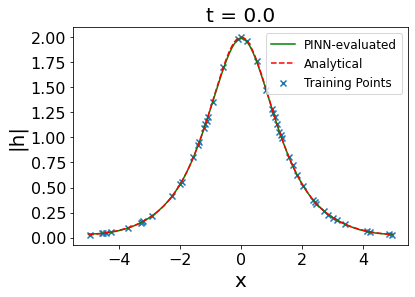}
  \label{H_t_0.0_0err}
  }
  \subfloat[]{
  \includegraphics[width=0.24\textwidth]{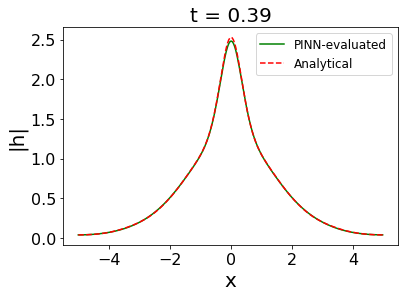}
  \label{H_t_0.39_0err}
  }
  \subfloat[]{
  \includegraphics[width=0.24\textwidth]{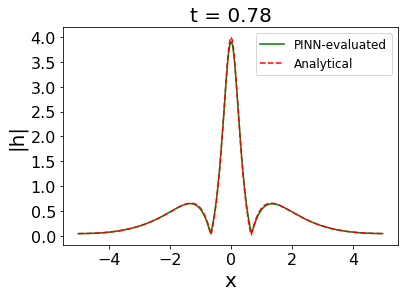}
  \label{H_t_0.78_0err}
  }
  \subfloat[]{
  \includegraphics[width=0.24\textwidth]{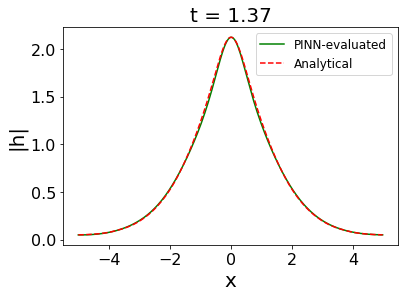}
  \label{H_t_1.37_0err}
  } \\
  \subfloat[]{
  \includegraphics[width=0.24\textwidth]{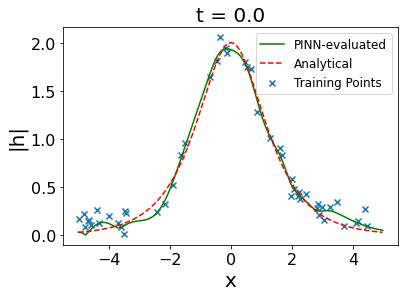}
  \label{H_t_0.0_01err}
  }
  \subfloat[]{
  \includegraphics[width=0.24\textwidth]{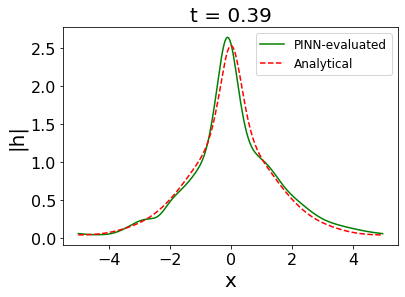}
  \label{H_t_0.39_01err}
  }
  \subfloat[]{
  \includegraphics[width=0.24\textwidth]{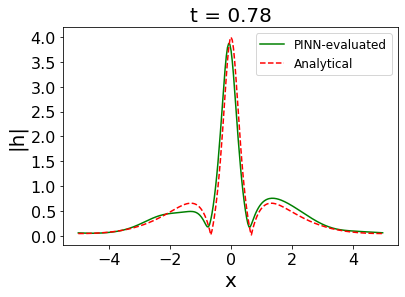}
  \label{H_t_0.78_01err}
  }
  \subfloat[]{
  \includegraphics[width=0.24\textwidth]{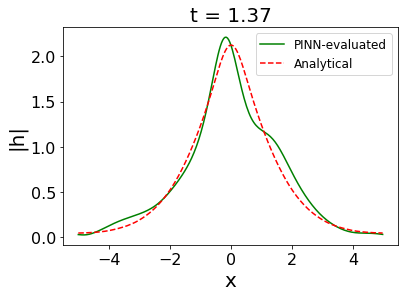}
  \label{H_t_1.37_01err}
  }
  \caption[]{The PINN-evaluated solution for $|h(x,t)|$ from the Schr\"odinger equation at different timeslices, $t = 0, 0.39, 0.78, 1.37$, from left to right for error-free boundary data (top) and corrupted boundary data with (bottom). The additive errors on the boundary data are taken independently from samples of zero mean Gaussian distribution with $\sigma = 0.1$.  The points marked with the blue cross (x) pointer in the leftmost set of plots indicate the samples on the initial timeslice used to train the PINN.}
  \label{fig:PINN-schrod}
\end{figure}

Figure~\ref{fig:PINN-schrod} shows the evolution of the complex field in the Schr\"odinger equation as evaluated by a vanilla PINN at four different timeslices, taken at \mbox{$t = 0, 0.39, 0.78, 1.37$}. The performance of the PINN for error-free data on the initial timeslice is shown in Figures~\ref{fig:PINN-schrod}(a)-(d). We use the same architecture to repeat the exercise while training on corrupted data on initial timeslice by letting $\Theta_u = \Theta_v = 1$ and additive errors generated by drawing samples from zero-mean Gaussian distribution with $\sigma = 0.1$. The performance of the PINN in evaluating the complex field magnitude at the same time instances is shown in Figures~\ref{fig:PINN-schrod}(e)-(h). The effect of introducing corrupted data on initial timeslice becomes evident when comparing Figures~\ref{fig:PINN-schrod}(a)-(d) with Figures~\ref{fig:PINN-schrod}(e)-(h). The PINN tends to overfit on the initial timeslice and eventually propagates these errors on the following timeslices.

\subsubsection{Burgers' Equation}

\begin{figure}
    \centering
    \subfloat[]{
        \includegraphics[width=0.24\textwidth]{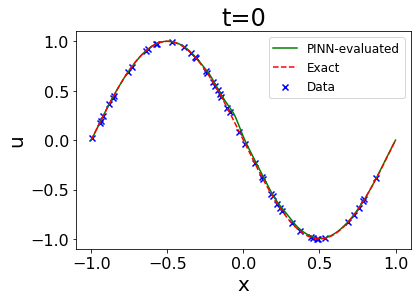}
    }
    \subfloat[]{
        \includegraphics[width=0.24\textwidth]{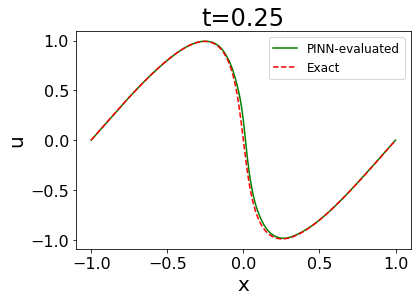}
    }
    \subfloat[]{
        \includegraphics[width=0.24\textwidth]{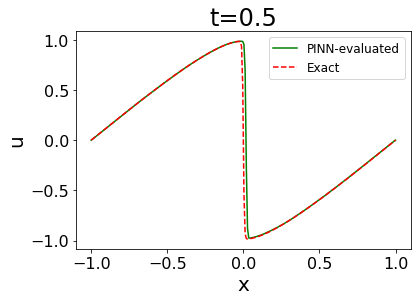}
    }
    \subfloat[]{
        \includegraphics[width=0.24\textwidth]{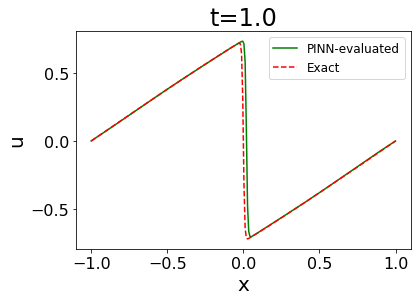}
    } \\
    \subfloat[]{
        \includegraphics[width=0.24\textwidth]{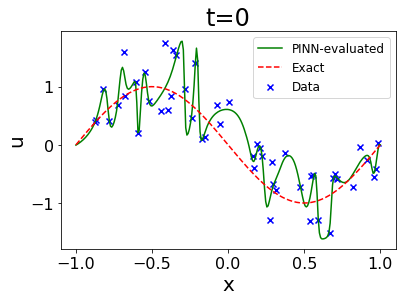}
    }
    \subfloat[]{
        \includegraphics[width=0.24\textwidth]{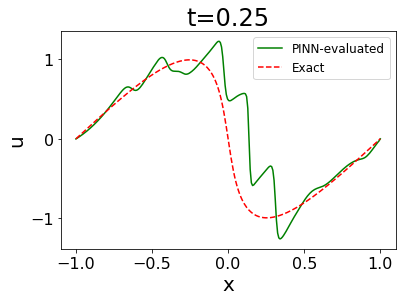}
    }
    \subfloat[]{
        \includegraphics[width=0.24\textwidth]{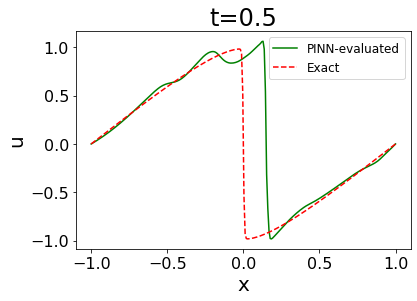}
    }
    \subfloat[]{
        \includegraphics[width=0.24\textwidth]{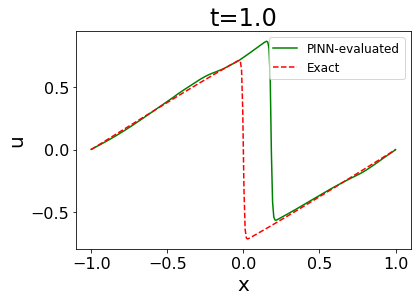}
    }
    
    \caption[]{The PINN-evaluated solution for $u(x,t)$ from the Burgers' equation at different timeslices, $t = 0, 0.25, 0.50, 1.00$, from left to right for error-free boundary data (top) and corrupted boundary data with (bottom). The additive errors on the boundary data are taken from independently samples of zero mean Gaussian distribution with $\sigma = 0.5$.  The points marked with the blue cross (x) pointer in the leftmost set of plots indicate the samples on the initial timeslice used to train the PINN.
    }
    \label{fig:PINN-burger}
\end{figure}

We consider the Burgers' equation in one spatial dimension with Dirchlet boundary conditions as a second example. Widely used in fluid dynamics and nonlinear acoustics, this nonlinear PDE has been widely studied as a benchmark example in the PINN literature~\cite{pinn-main,pinn-adversarial}. The PDE and the boundary conditions for 1D Burgers' equation are given as :
\begin{align}
    &\partialD{u}{t} + u \partialD{u}{x} = \nu \partialD{^2 u}{x^2} \label{eqn:pde-burger} \\
    &u(-1,t) = u(1,t) = 0 \label{eqn:burger-bc}\\
    &u(x,0) = -\sin(\pi x) + \Theta_u\epsilon^u \label{eqn:burger-ic}
\end{align}
where the domain boundary is given as $(x,t) \in [-1, 1] \times [0,1]$. To set the benchmark for this probelm, we solve Eqn.~\ref{eqn:pde-burger} subject to boundary conditions in Eqns.~\ref{eqn:burger-bc} and ~\ref{eqn:burger-ic} with $\nu = \frac{0.01}{\pi}$ using a PINN without noise ($\Theta_u = 0$) in the initial data. We used an MLP with 4 hidden layers, each with 40 nodes. We trained with $N_c = 10000$ collocation points to enforce the physics (Eqn.~\ref{eqn:pde-burger}) and $N_b = 150$ points on the boundary, 50 points for enforcing the initial condition in Eqn~\ref{eqn:burger-ic} and 50 points on each of the spatial boundaries at $x= -1, 1$ to enforce each of the Dirichlet conditions in Eqn.~~\ref{eqn:burger-bc}. The loss function is constructed according to Eq. \ref{eqn:loss-pinn} with $\alpha_{()} = 1.0$. Figure~\ref{fig:PINN-burger}(a)-(d) shows the evolution of the field $u(x,t)$ in the Burgers' equation as evaluated by a vanilla PINN at four different timeslices, taken at \mbox{$t = 0, 0.25, 0.50, 1.0$}.

Next, we repeat the exercise for Burgers' equation by introducing additive corruption with $\Theta_u = 1$ and $\sigma = 0.5$. Figures~\ref{fig:PINN-burger}(e)-(h) show the corresponding line shapes of the Burgers' field for the aforementioned timestamps. Similar to what was found for the Schr\"odinger Equation, the vanilla PINN architecture fails to auto-correct for the corruption in initial data and ends up overfitting on the initial timeslice and eventually propagates these errors to later timeslices.

\subsection{Regularization of PINNs}

Both Figures~\ref{fig:PINN-schrod} and \ref{fig:PINN-burger} show a PINN's inherent inability to self-correct when trained with error-corrupted data on the initial timeslice. The PINN rather learns to overfit on the initial timeslice and eventually propagates the initial errors to later timeslices. This propagation of error to later timeslices is a direct consequence of having no regularization in the loss term in Eqn.~\ref{eqn:loss-pinn} to constrain overfitting at the domain boundary. 

In principle, this is not very different from overfitting in classic regression problems with a high degree polynomial or introduction of bias in an un-regularized regression by outliers. This naively indicates that additional regularization might be useful to limit the propagation of errors. However, our investigations indicates that some of the most physically intuitive choices for regularizing constraints have little impact on error propagation and boundary overfitting. 

We consider two unique choices of regularizers. First, inspired by the cPINN architecture,  we impose the constraint of functional and flux continuity at arbitrary spatio-temporal boundaries to constrain the PDE solution. Second, we explore the possibility of using physical conservation laws as additional sources of regularization. In the following subsections, we explore the impact of using such regularization schemes in training PINNs with corrupted boundary data.

\subsubsection{Functional and flux continuity at subdomain interfaces}
\label{subsec:cpinn}

The cPINN architecture inspires a useful regularization that imposes continuity of the field and its flux across domain boundaries. From a physics standpoint, these regularizations can be thought of as additional conservation laws that ensures continuity and differentiability of a field across subdomains.  In this subsection, we explore the impact of including this term in the training loss function in Eqn.~\ref{eqn:loss-cpinn} in controlling propagation of uncorrelated errors at sampling points on the initial timeslice.

\begin{figure}
\centering
  \subfloat[]{
  \includegraphics[width=0.24\textwidth]{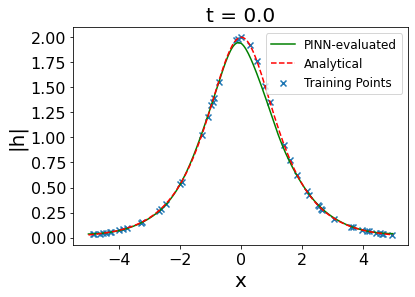}
  \label{H_t_0.0}
  }
  \subfloat[]{
  \includegraphics[width=0.24\textwidth]{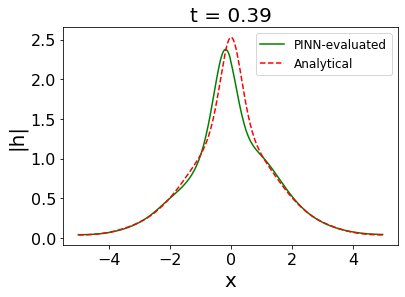}
  \label{H_t_0.39}
  }
  \subfloat[]{
  \includegraphics[width=0.24\textwidth]{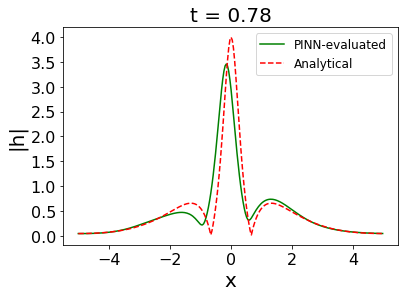}
  \label{H_t_0.78}
  }
  \subfloat[]{
  \includegraphics[width=0.24\textwidth]{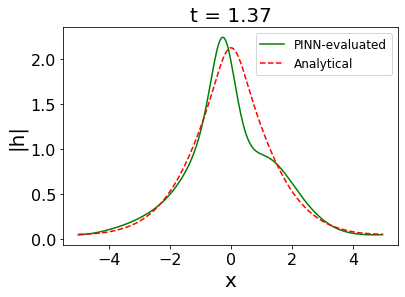}
  \label{H_t_1.37}
  } \\
  \subfloat[]{
  \includegraphics[width=0.24\textwidth]{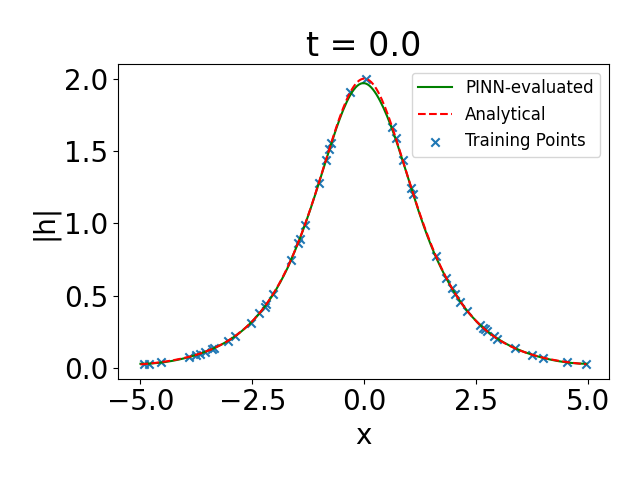}
  \label{H_t_0.0}
  }
  \subfloat[]{
  \includegraphics[width=0.24\textwidth]{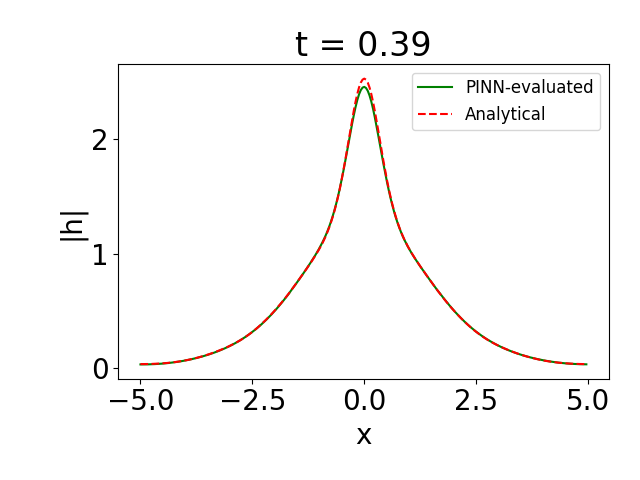}
  \label{H_t_0.39}
  }
  \subfloat[]{
  \includegraphics[width=0.24\textwidth]{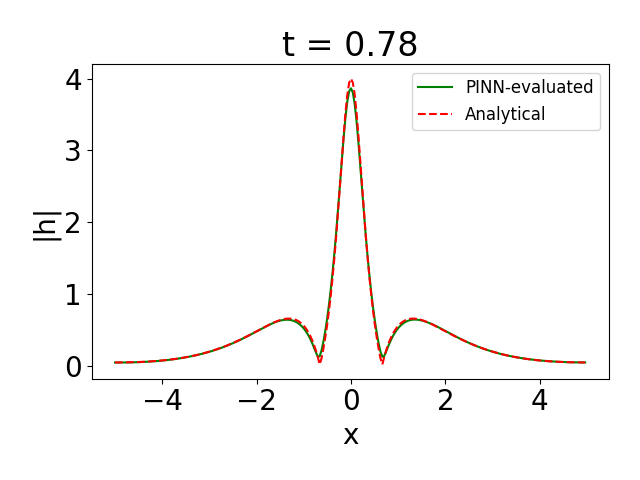}
  \label{H_t_0.78}
  }
  \subfloat[]{
  \includegraphics[width=0.24\textwidth]{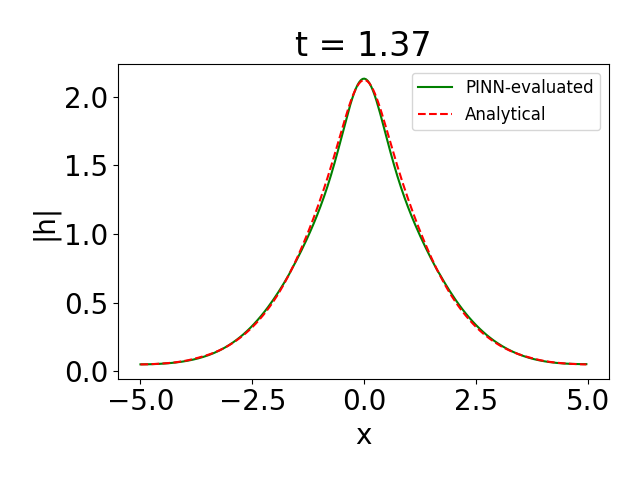}
  \label{H_t_1.37}
  }
  \caption[]{The cPINN-evaluated $|h(x,t)|$  for two (top row) and three (bottom row) equal subdomains at different timeslices, $t = 0, 0.39, 0.78, 1.37$, from left to right when no error is introduced on initial time-slice. The points marked with the blue cross (x) pointer in the leftmost set of plots indicate the samples on the initial timeslice used to train the cPINN. }
  \label{fig:cPINN23-schrod-noerror}
\end{figure}

\begin{figure}
\centering
 \subfloat[]{
  \includegraphics[width=0.33\textwidth]{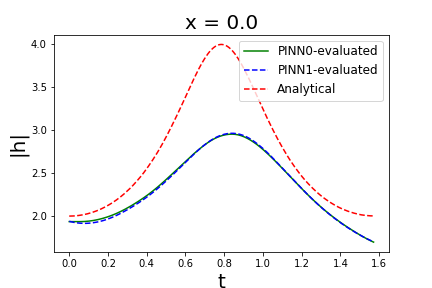}
  \label{70nodes_U_t_0.0}
  }
  \subfloat[]{
  \includegraphics[width=0.33\textwidth]{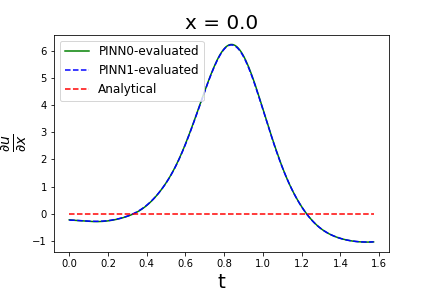}
  \label{70nodes_U_t_0.0}
  }
  \subfloat[]{
  \includegraphics[width=0.33\textwidth]{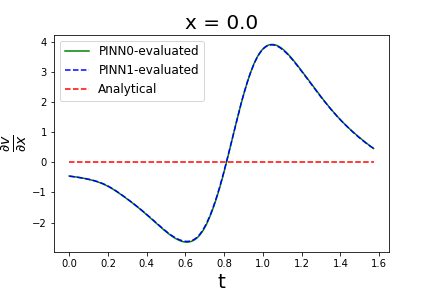}
  \label{70nodes_U_t_0.0}
  }
  
  \caption[]{ The cPINN evaluated lineshape  for $|h(x,t)|$ (left), $\partialD{u}{x}$ (middle), and $\partialD{v}{x}$ (right) at the subdomain interface $(x=0)$ for a two subdomain cPINN as a function of $t$. \texttt{PINN0 (PINN1)} refers to the PINN trained to solve the PDE on a spatial boundary of $-5 \le x \le 0 \quad (0 \le x \le 5)$.}
  \label{fig:cPINN-localminima}
\end{figure}

One immediate concern is that the convergence of the cPINN is susceptible to exact choices of how many subdomains are chosen and where those boundaries are located. Based on the location of the subdomain boundaries, the cPINN's capacity to converge to the global minimum of the training loss function, can be significantly impacted. 
To illustrate this, we compare the performance of cPINNs in solving the Schr\"odinger equation with two and three equal spatial subdomains trained on error-free boundary data. The results are shown in Figure~\ref{fig:cPINN23-schrod-noerror}. Evidently, a three subdomain cPINN better recovers the analytical solution. However, the failure of a two subdomain cPINN, to capture the solution of a PDE even for error-free boundary data is intriguing. This observation yields a deeper insight to the impact of adding additional regularizers with the PINN loss function in Eqn.~\ref{eqn:loss-pinn}. As can be seen in Figure~\ref{fig:cPINN23-schrod-noerror}(a), the two subdomain cPINN moderately deviates from the analytical solution on the initial timeslice at the expense of converging to the local minima introduced by the inclusion of the interface loss. Figure~\ref{fig:cPINN-localminima} shows how the PINNs trained on different subdomains converge to identical functional and flux values at the subdomain boundary but eventually experiences large deviations from the analytical solution which requires $\partialD{u}{x}(x,0) = \partialD{v}{x}(x,0)  = 0$. 
It is apparent that choice of subdomain boundary at $x=0$ plays an important role in causing such deviating solutions. As the real and imaginary fields reach local extrema at $x=0$ for all values of $t$, even small deviations in estimating the local gradients at the interface can create a cascading effect in the evaluation of the complex field across subdomain boundaries as we can see in Figures~\ref{fig:cPINN-localminima}.

\begin{figure}
\centering
  \subfloat[]{
  \includegraphics[width=0.24\textwidth]{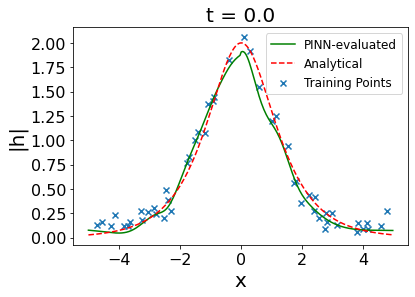}
  \label{H_t_0.0}
  }
  \subfloat[]{
  \includegraphics[width=0.24\textwidth]{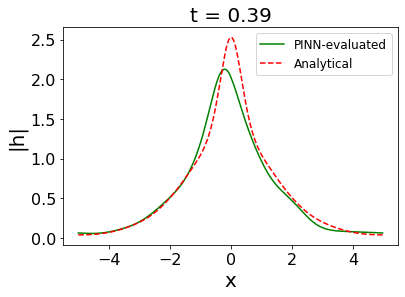}
  \label{H_t_0.39}
  }
  \subfloat[]{
  \includegraphics[width=0.24\textwidth]{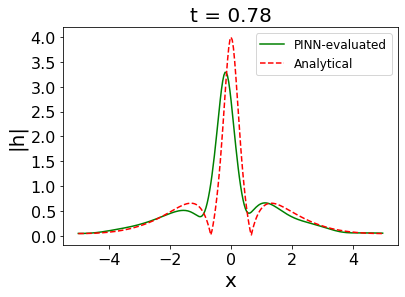}
  \label{H_t_0.78}
  }
  \subfloat[]{
  \includegraphics[width=0.24\textwidth]{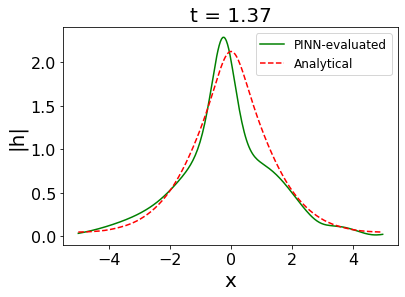}
  \label{H_t_1.37}
  } \\
  \subfloat[]{
  \includegraphics[width=0.24\textwidth]{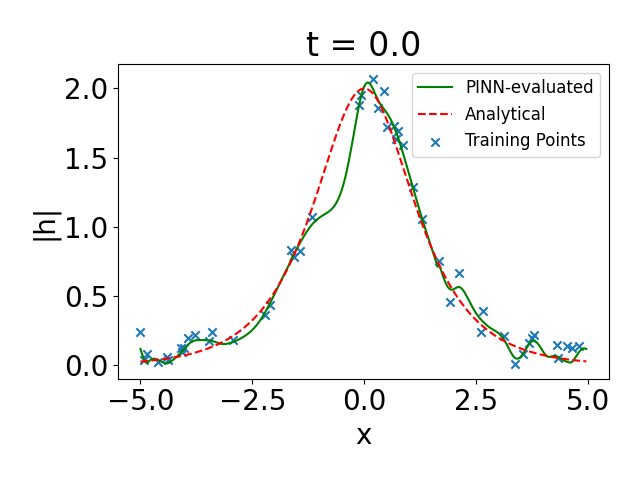}
  \label{H_t_0.0}
  }
  \subfloat[]{
  \includegraphics[width=0.24\textwidth]{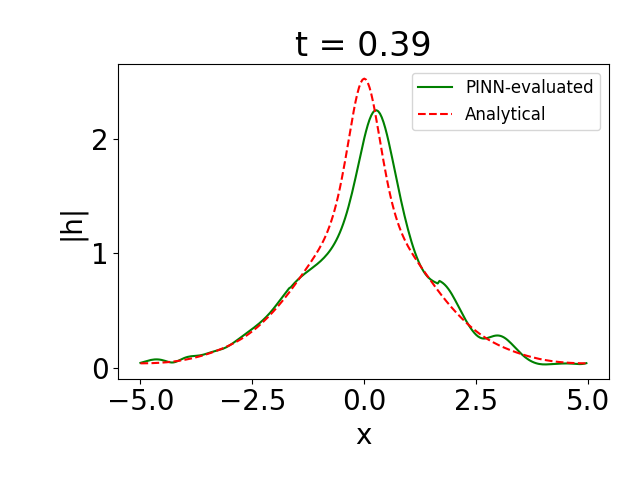}
  \label{H_t_0.39}
  }
  \subfloat[]{
  \includegraphics[width=0.24\textwidth]{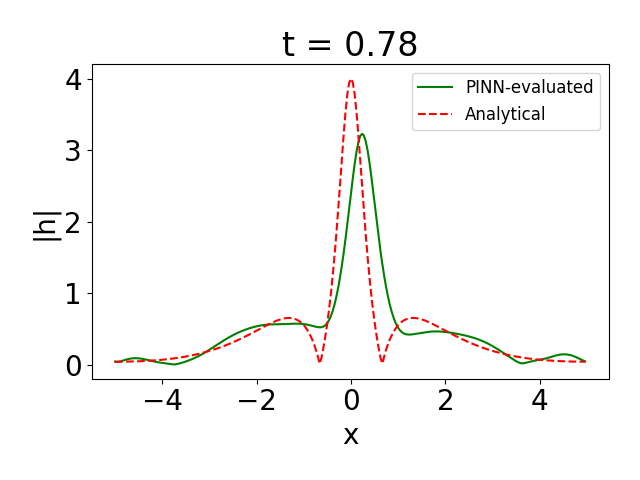}
  \label{H_t_0.78}
  }
  \subfloat[]{
  \includegraphics[width=0.24\textwidth]{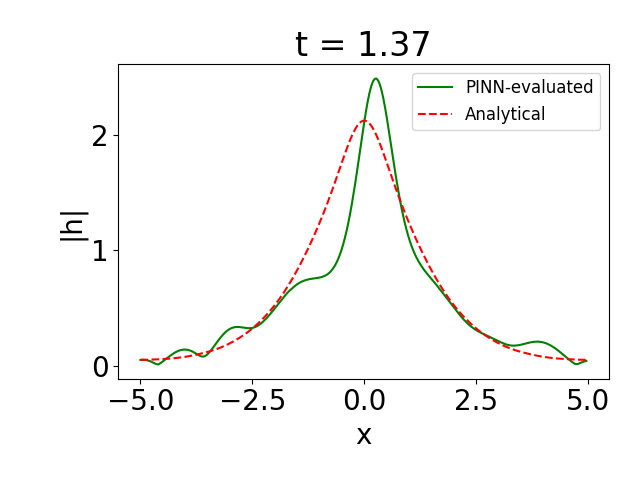}
  \label{H_t_1.37}
  }
  \caption[]{The cPINN-evaluated $|h(x,t)|$  for two (top row) and three (bottom row) subdomains at different timeslices, $t = 0, 0.39, 0.78, 1.37$, from left to right when additive Gaussian errors with $\sigma = 0.1$ is introduced on initial time-slice. The points marked with the blue cross (x) pointer in the leftmost set of plots indicate the samples on the initial timeslice used to train the cPINN. }
  \label{PINN-error-23layer}
\end{figure}

The result of such instabilities in convergence, is evidential consequence of such  cPINN-inspired regularizations and the architecture's failure to be robust to such noisy perturbations. As also shown in Figure~\ref{PINN-error-23layer}, regularization of functional and flux continuity at subdomain boundaries does not provide the necessary safeguard to ensure robustness.

\begin{figure}[!h]
    \centering
    
    \subfloat[]{
        \includegraphics[width=0.24\textwidth]{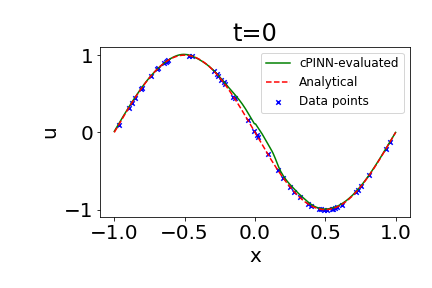}
    }
    \subfloat[]{
        \includegraphics[width=0.24\textwidth]{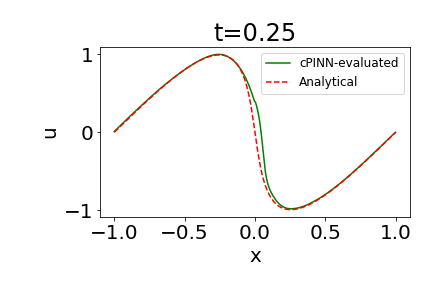}
    }
    \subfloat[]{
        \includegraphics[width=0.24\textwidth]{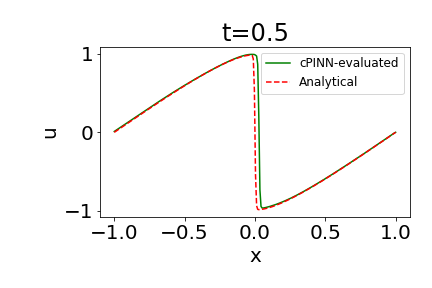}
    }
    \subfloat[]{
        \includegraphics[width=0.24\textwidth]{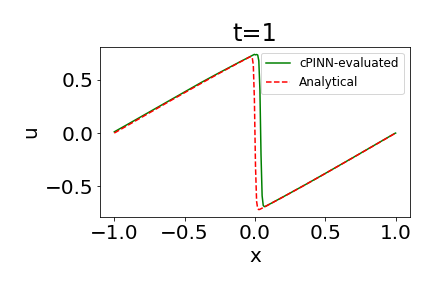}
    }
    \\
    
    \subfloat[]{
        \includegraphics[width=0.24\textwidth]{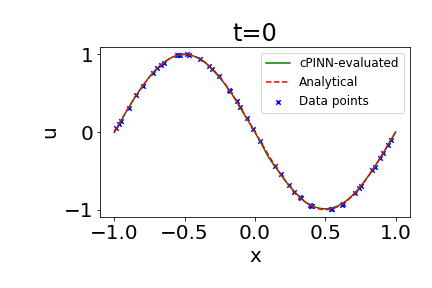}
    }
    \subfloat[]{
        \includegraphics[width=0.24\textwidth]{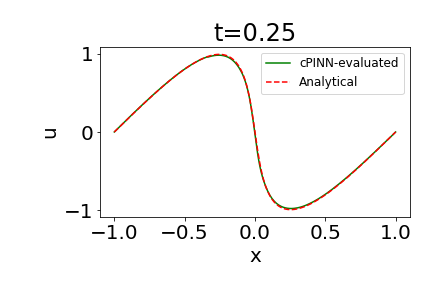}
    }
    \subfloat[]{
        \includegraphics[width=0.24\textwidth]{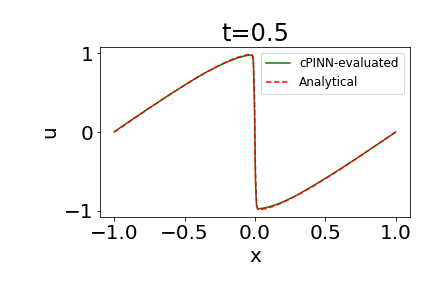}
    }
    \subfloat[]{
        \includegraphics[width=0.24\textwidth]{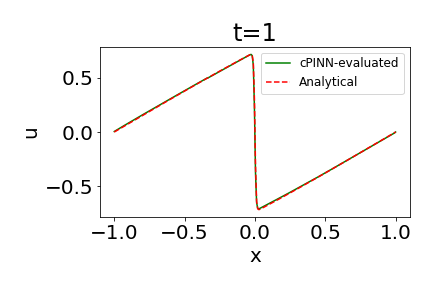}
    }
    \\
    \subfloat[]{
        \includegraphics[width = 0.24\textwidth]{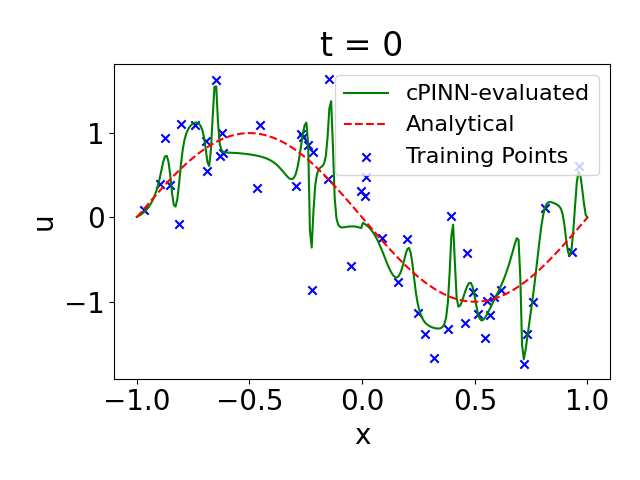}
    }
    \subfloat[]{
        \includegraphics[width = 0.24\textwidth]{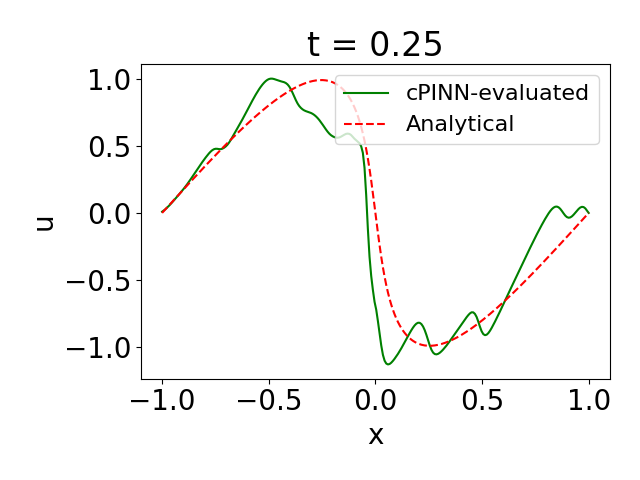}
    }
    \subfloat[]{
        \includegraphics[width = 0.24\textwidth]{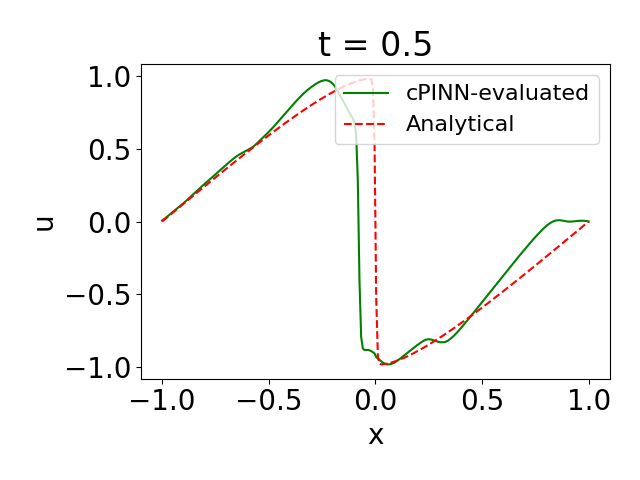}
    }
    \subfloat[]{
        \includegraphics[width = 0.24\textwidth]{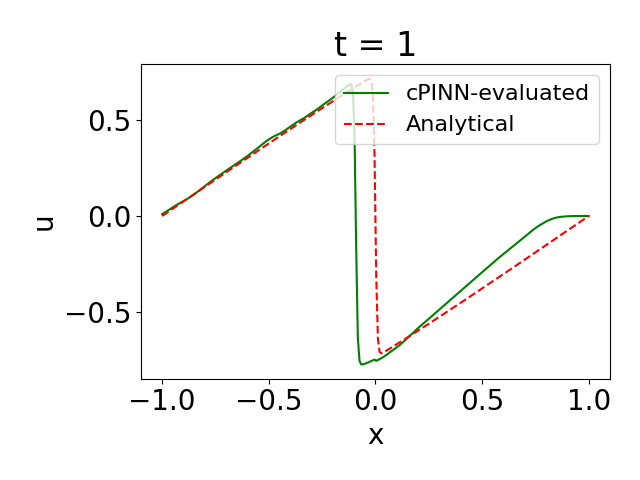}
    }
    \\
    \subfloat[]{
        \includegraphics[width = 0.24\textwidth]{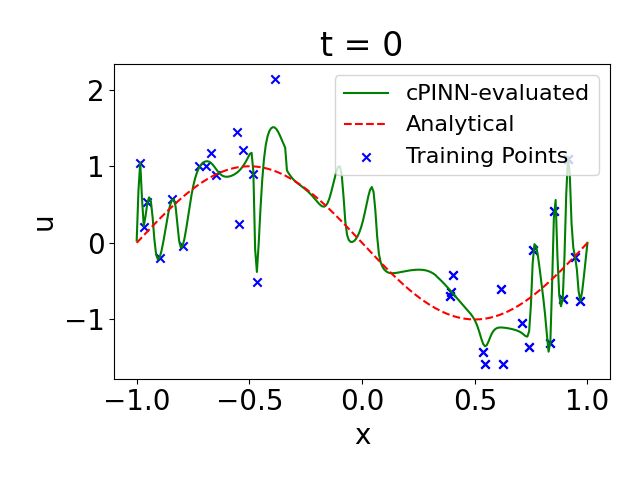}
    }
    \subfloat[]{
        \includegraphics[width = 0.24\textwidth]{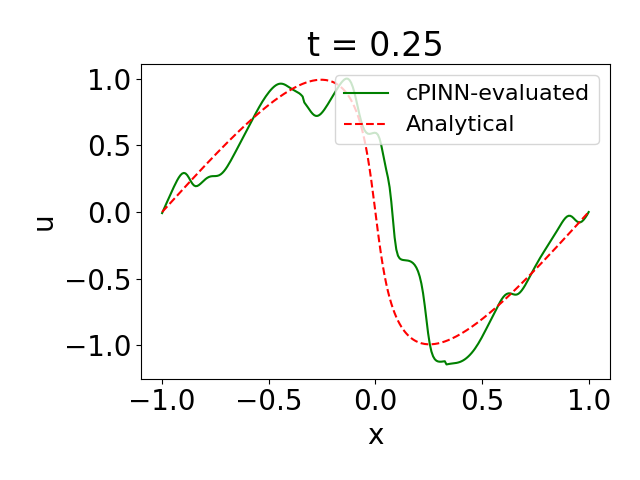}
    }
    \subfloat[]{
        \includegraphics[width = 0.24\textwidth]{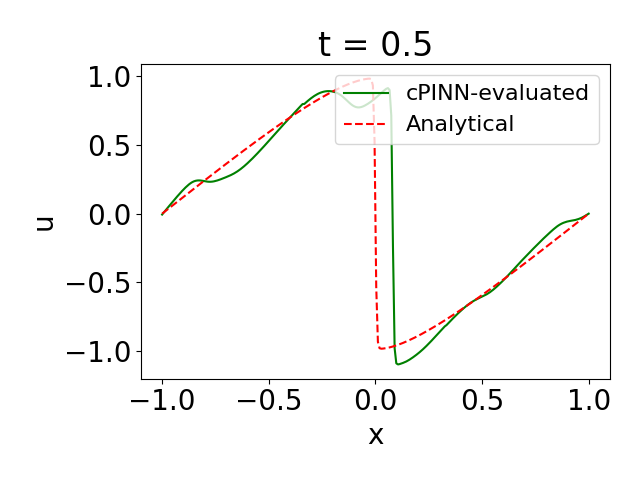}
    }
    \subfloat[]{
        \includegraphics[width = 0.24\textwidth]{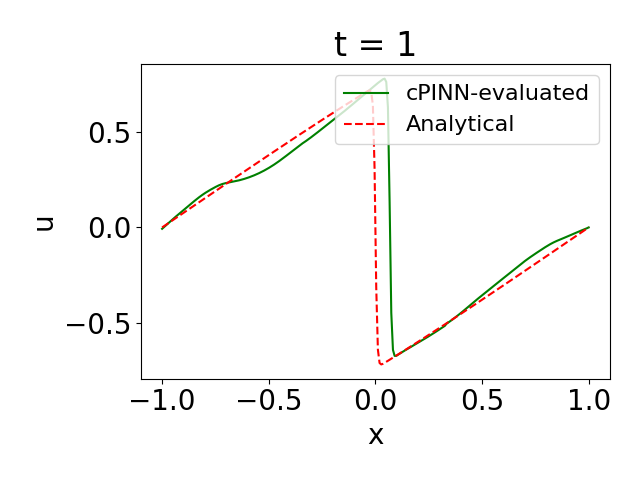}
    }

    \caption{cPINN-evaluated solution to Burgers' equation with (a-d) two subdomains with no error, (e-h) three subdomains with no error, (i-l) two subdomains with error, and (m-p) three subdomains with error. The error on individual datapoints, whenever applied, have been taken from additive zero-mean Gaussian distribution with $\sigma = 0.5$.
    }
    \label{fig:cPINN_Burgers}
\end{figure}

When we repeat the same set of exercises for the Burgers' equation, we see very similar results, as shown in Figures~\ref{fig:cPINN_Burgers}. Similar to what we have seen for the Schr\"odinger equation, placing the subdomain boundaries at functionally critical points can destabilize and deteriorate the quality of the solution learned by the deep network. It can be seen in Figures~\ref{fig:cPINN_Burgers}(a)-(d) from the deviation of the Burgers' field's   predicted behavior at later timeslices even without any error introduced on the initial timeslice when two subdomains are considered with an interface at $x = 0$. However, the function is almost identically recovered when the data is perfect on the initial timelice with three subdomains as shown in Figures~\ref{fig:cPINN_Burgers}(e)-(h). This tendency of a PINN-like architecture to converge to a local minima instead of the global minima almost infallibly deteriorates the quality of convergence when error is introduced on the initial timeslice, and thus the solution departs significantly from its ideal behavior. This is apparent in both the  two subdomain  (Figures~\ref{fig:cPINN_Burgers}(i)-(l)) and the  three subdomain cPINNs (Figures~\ref{fig:cPINN_Burgers}(m)-(p)).

\subsubsection{Conservation law constraints}
\label{subsec:conserv-law}

Physical laws are often subject to a number of conservation laws. While in many cases these conservation laws emerge as direct consequences of the mathematical structure of the PDE, explicitly enforcing such conservation laws will back-propagate additional constraining gradients for the NN hyperparameters.  We can explicitly include these conservation laws into the loss function.

\begin{figure}[!h]
    \centering
    \subfloat[]{
        \includegraphics[width = 0.3\textwidth]{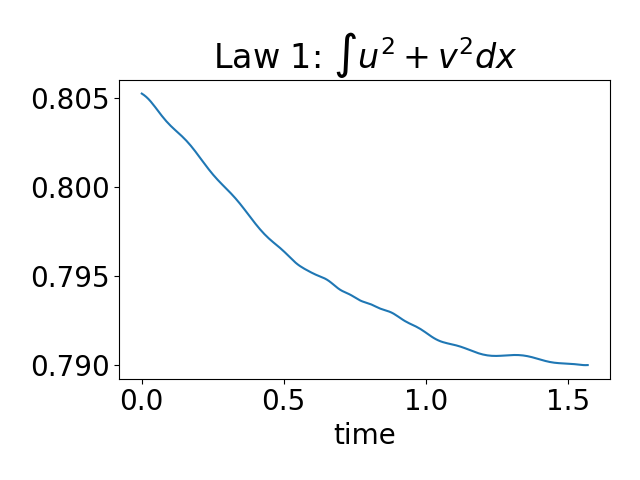}
    }
    \subfloat[]{
        \includegraphics[width = 0.3\textwidth]{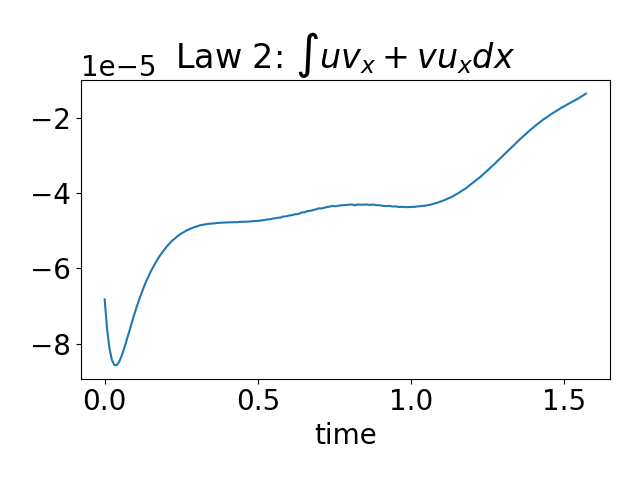}
    }
    \subfloat[]{
        \includegraphics[width = 0.34\textwidth]{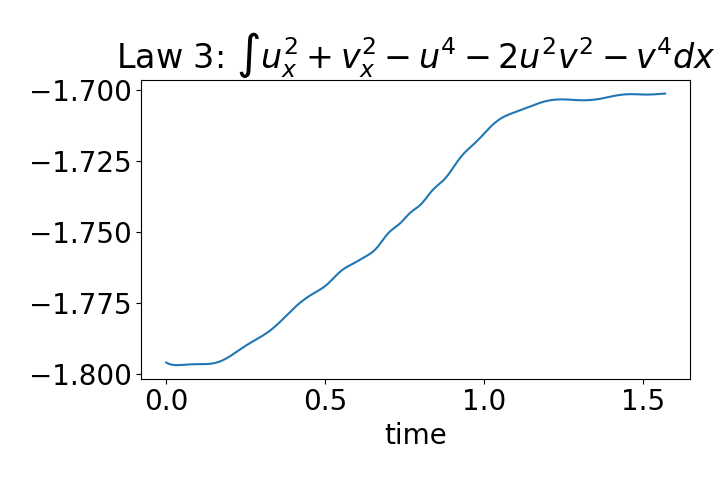}
    }
    \\
    \subfloat[]{
        \includegraphics[width = 0.3\textwidth]{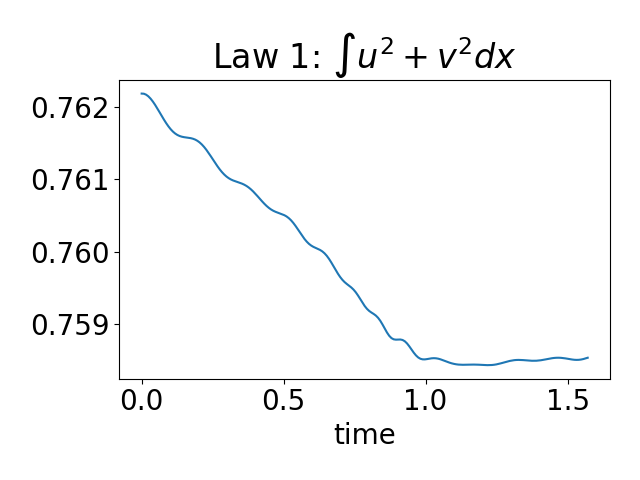}
    }
    \subfloat[]{
        \includegraphics[width = 0.3\textwidth]{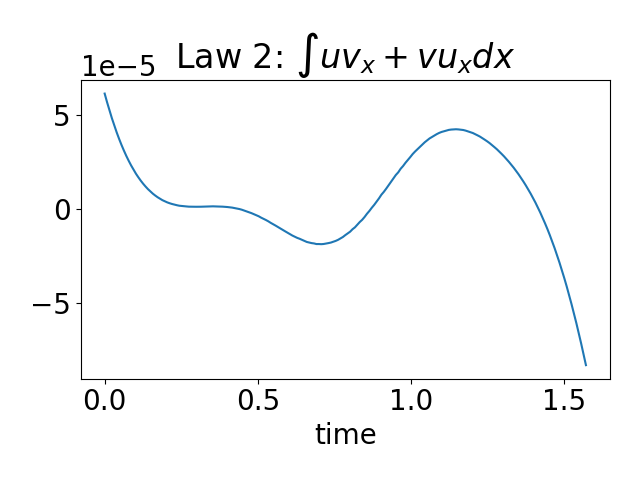}
    }
    \subfloat[]{
        \includegraphics[width = 0.34\textwidth]{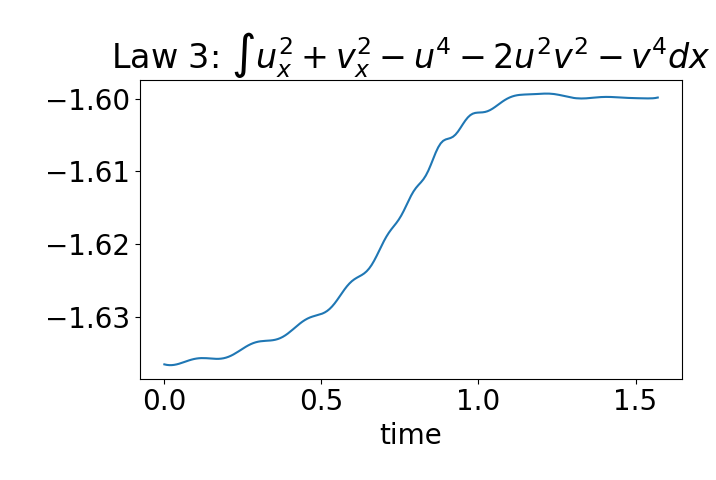}
    }
    \caption{Plots of three of the conservation laws (Eqns.~\ref{eqn:cons-prob}-\ref{eqn:cons-3}) of the nonlinear Schr\"odinger equation for (a-c) a vanilla PINN  and (d-f) a PINN constrained by the first conservation law. The PINNs were trained with initial data corrupted with $\sigma=0.1$ Gaussian noise. The constrained PINN has better performance of conservation laws 1 and 3, while both PINNs satisfy conservation law 2 nearly exactly.}
    \label{fig:conservation}
\end{figure}

\begin{figure}[!h]
    \centering
    \subfloat[]{
        \includegraphics[width = 0.24\textwidth]{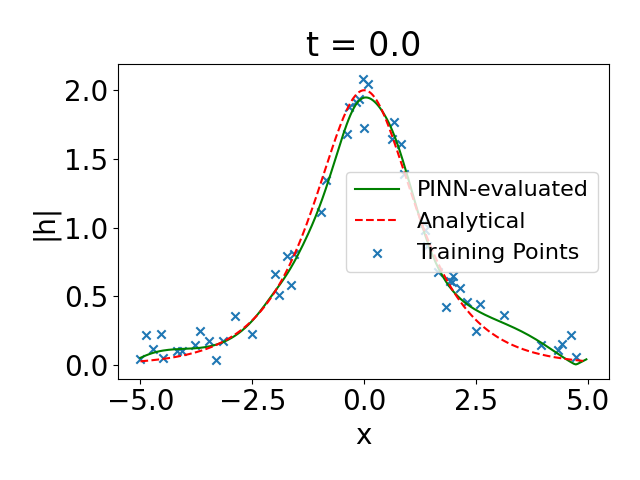}
    }
    \subfloat[]{
        \includegraphics[width = 0.24\textwidth]{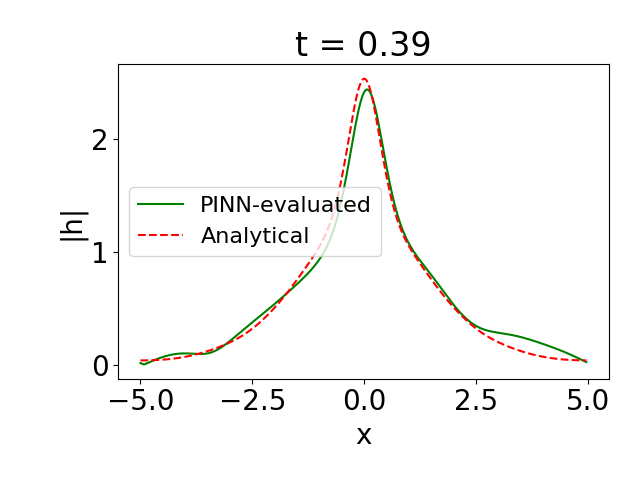}
    }
    \subfloat[]{
        \includegraphics[width = 0.24\textwidth]{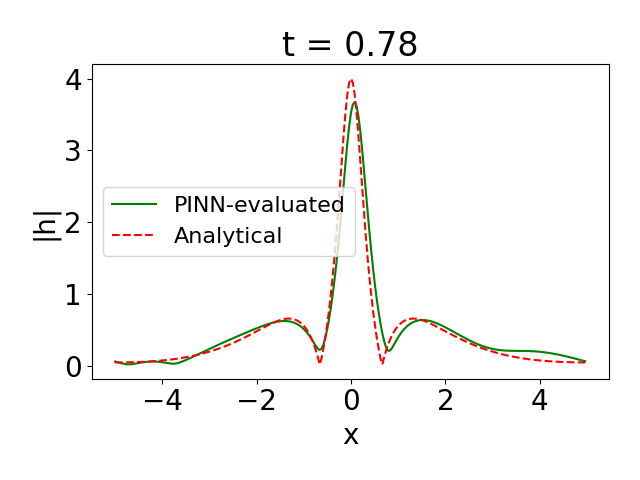}
    }
    \subfloat[]{
        \includegraphics[width = 0.24\textwidth]{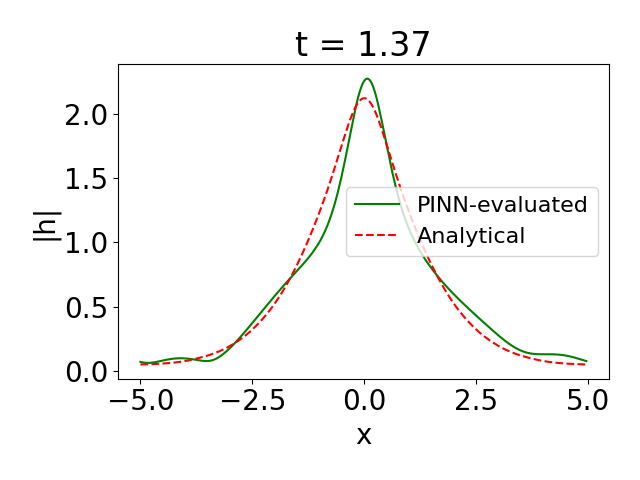}
    }
    \caption{The PINN-evaluated $|h(x,t)|$ at different timeslices, $t = 0, 0.39, 0.78, 1.37$, from left to right. In this case the PINN is constained by the Schrodinger equations first conservation law: $\frac{d}{dt}\int |h|^2 dx=0$. The training data on the initial timeslice is subject to measurement errors, modeled by a Gaussian random variable with zero mean and a standard deviation of $\sigma = 0.1$. The points marked with the blue cross (x) pointer in the leftmost set of plots indicate the samples on the initial timeslice used to train the PINN.
    }
    \label{fig:cons-law-PINN}
\end{figure}

For example, one of the major consequences of non-linear Schr\"{o}dinger equation is global conservation of the squared absolute value of the complex Schr\"{o}dinger field $|h(x,t)|^2$ 
\begin{equation}
    \int |h(x,t)|^2 dx = \int (u(x,t)^2 + v(x,t)^2) dx = C
    \label{eqn:cons-prob}
\end{equation}
where $C$ is a constant. A number of other conserved quantities follow for the 1D nonlinear Schr\"{o}dinger equation we are considering \cite{DeriveConsLawsNLS, MethodConsLawsNLS, LocalConsLawsNLS}, which include:
\begin{align}
    & \int \left( u\partialD{v}{x} + v \partialD{u}{x} \right) dx \label{eqn:cons-2} \\
    & \int \left( \left| \partialD{h}{x} \right|^2 - |h|^4 \right) dx \label{eqn:cons-3}
\end{align}
We can constrain the solution by explicitly including these conservation laws as regularizers in the loss function. For instance, the probability conservation law in Eqn.~\ref{eqn:cons-prob}, along with the requirement of probability confinement within the spatially bounded region for the domain of Eqn.~\ref{eqn:pde-schrod} requires that
\begin{equation}
    0 = \frac{d}{dt} \int_a^b |h|^2 dx = \frac{d}{dt} \int_a^b u^2 + v^2 dx = 2\int_a^b uu_t + vv_t dx
\end{equation}
where $(a,b)$ is the spatial domain. We can approximate this integral using time-sliced collocation points:
\begin{equation}
    \int_a^b uu_t + vv_t dx \approx \frac{b-a}{n_{c,t}} \sum_{i=1}^{n_{c,t}} u(x_i,t) u_t(x_i,t)  + v(x_i,t)v_t(x_i,t) 
\end{equation}
where $n_{c,t}$ represents the number of collocation points chosen over the spatial subdomain at timeslice $t$. We can define the conservation loss to be
\begin{equation}
    {\mathcal{L}}_C=\frac{1}{N_{t}}\sum_{i=0}^{N_{t}}\frac{1}{n_{c,t}}\sum_{j=0}^{n_{c,t}}\left({u\left(x_j,t_i\right)u_t\left(x_j,t_i\right)+v\left(x_j,t_i\right)v_t\left(x_j,t_i\right)}\right)^2\ \
\end{equation}
We train a PINN with this conservative constraint on data appended with the loss function in Eqn.~\ref{eqn:loss-pinn}.
As seen in Figure~\ref{fig:conservation}, the constrained PINN better represents the conservation laws, with the overall range for dispersion of cumulative probability effectively reduced with the inclusion of conservative constraints in the PINN loss function. However, the observed evolution of the field in Figure~\ref{fig:cons-law-PINN}, obtained by applying this constraint does not result in superior accuracy. Errors are still propagated from the initial timeslice throughout the spatio-temporal domain.

As a second example of the implication of conservation laws as regularizers, we take the example of Cole-Hopf transformation~\cite{cole,hopf} for the Burgers' equation. Based on proper mathematical wisdom developing analytical solutions to PDEs, methods like the Cole-Hopf transform have been found useful to convert one family of PDEs into another whose analytical solution is known and rather simple to compute. The Cole-Hopf transformation converts the nonlinear Burgers' equation to a linear Heat equation. This transformation is defined by making the following change of variables:

\begin{equation}
    u(x,t) = -2\nu\partialD{v}{x}(x,t)
    \label{eqn:Cole-Hopf}
\end{equation}

The transformed field $v(x,t)$ satisfies the heat equation $\partial_t v = \nu \partial_{xx} v$. The viscous case of the Burgers' equation is for $\nu > 0$ causing a non-linear dissipative shock for small values of $\nu$. The inviscid case yields the equation having a non-linear hyperbolic conservation law. For our purpose, the conservation law for the modified field can be viewed as a regularizer to the conventional PINN loss function in Eqn.~\ref{eqn:loss-pinn}. 

\begin{equation}
    \mathcal{L}_{\mathrm{CH}} = \sum_{i,j} \left( v_t(x_i, t_j) - \nu v_{xx}(x_i, t_j) \right)^2
    \label{eqn:loss-CH}
\end{equation}

\begin{figure}[!h]
    \centering
    
    \subfloat[]{
        \includegraphics[width=0.24\textwidth]{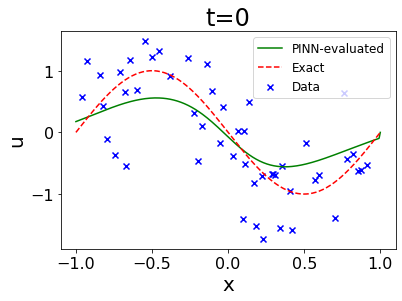}
    }
    \subfloat[]{
        \includegraphics[width=0.24\textwidth]{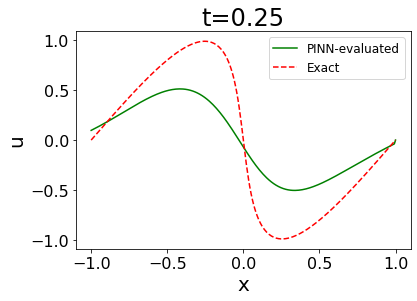}
    }
    \subfloat[]{
        \includegraphics[width=0.24\textwidth]{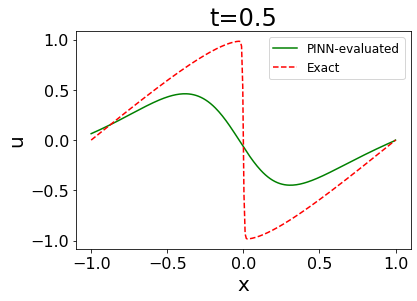}
    }
    \subfloat[]{
        \includegraphics[width=0.24\textwidth]{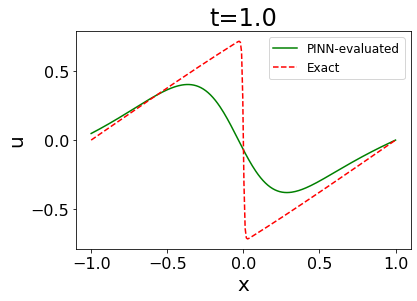}
    } \\
      \subfloat[]{
        \includegraphics[width=0.24\textwidth]{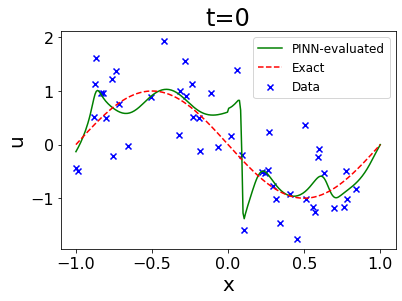}
    }
    \subfloat[]{
        \includegraphics[width=0.24\textwidth]{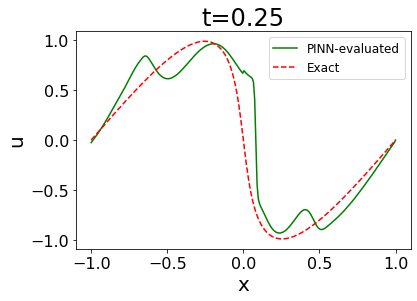}
    }
    \subfloat[]{
        \includegraphics[width=0.24\textwidth]{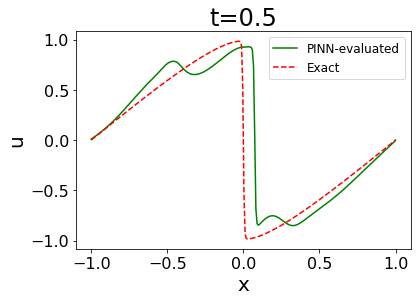}
    }
    \subfloat[]{
        \includegraphics[width=0.24\textwidth]{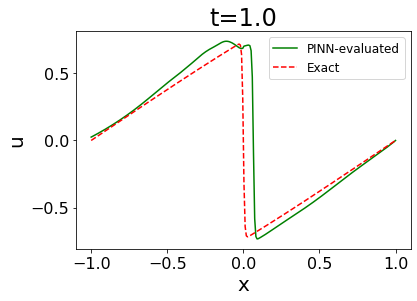}
    }
    \\
      \subfloat[]{
        \includegraphics[width=0.24\textwidth]{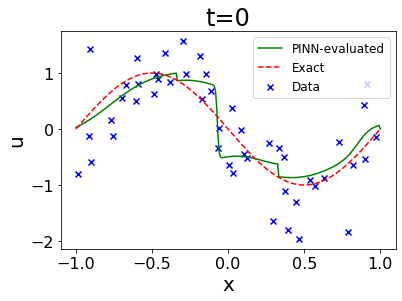}
    }
    \subfloat[]{
        \includegraphics[width=0.24\textwidth]{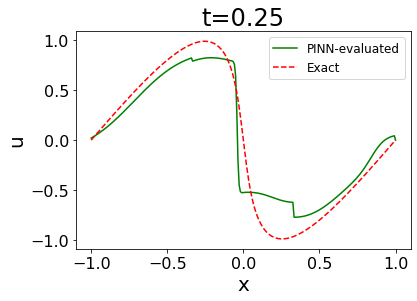}
    }
    \subfloat[]{
        \includegraphics[width=0.24\textwidth]{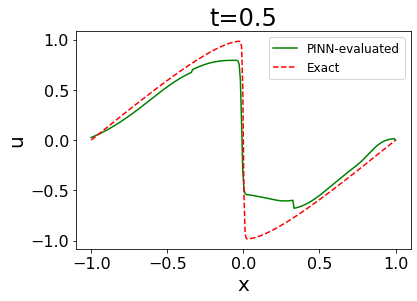}
    }
    \subfloat[]{
        \includegraphics[width=0.24\textwidth]{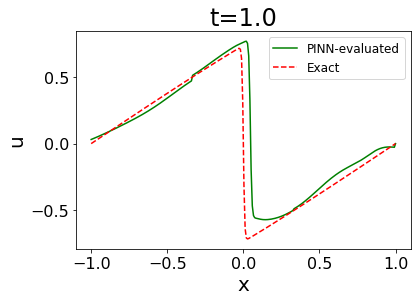}
    }
    
      \caption[]{The PINN-evaluated $u(x,t)$  at different timeslices, $t = 0, 0.25, 0.5, 1.0$, from left to right when additive Gaussian errors with $\sigma = 0.5$ is introduced on initial time-slice for when additional regularization in terms of Cole-Hopf transformation is added to the loss function. A single domain PINN used for the top row while the middle and bottom rows show result from 2 and 3 domain cPINNs respectively.}
  \label{fig:PINN-burger-CH}
 
    \end{figure}

 In addition to using the Cole-Hopf loss term as a regularizer for a vanilla PINN architecture, we additionally consider including the continuity criteria for equally split two and three subdomains. Figure~\ref{fig:PINN-burger-CH}(a)-(d) shows the time evolution of the Burgers' field $u(x,t)$ when the PINN is trained with the loss function including the Cole-Hopf term in Eqn.~\ref{eqn:loss-CH}. The functional approximation obtained from the PINN is much smoother compared to what we have observed in our previous exploration of regularized evaluation of the solution to the Burgers' equation (Figures~\ref{fig:PINN-burger} and \ref{fig:cPINN_Burgers}). It can be directly traced back to the fact that the Cole-Hopf transformed field is indeed an anti-derivative of the Burgers' field and in the neural architecture when implemented as a discrete integration acts as a smoothing operation in somewhat canceling out the effect of the error. The smoothing operation however eventually leads to underfitting and the evolution of the field at later timeslices is affected in a similar fashion.

Figure~\ref{fig:PINN-burger-CH}(e)-(h) [(i)-(l)] shows the results when we trained the Cole-Hopf regularized PINN with functional and flux continuity imposed at the interface of two [three] sub-domains. These results establish the inadequacy of the Cole-Hopf regularizer in establishing the robustness of the neural architecture and reinforces their tendency to converge to local minimum instead of reaching the intended global minimum. 

%% file: tex-paper/4GPSmoothing.tex
\section{Gaussian Process (GP) based Error Correction for PINNs}
\label{sec:gp-pinn}
As the previous section illustrates, physics-inspired regularization alone does not eliminate propagation of errors in PINNs. In fact, application of such constraints forces the PINN to converge to a local minimum that satisfies the physics of conservation laws for overfitted boundary conditions and eventually propagates the overfitting across the spatio-temporal domain. In this section, we seek the explore alternate solution to this problem by using smoothing techniques that safeguard the quality of fit by cross-validated regulation of smoothed boundary data. The model of corruption considered in the PDE model for 1D Schr\"odinger and Burgers' equations is a ubiquitous approximation for many physics processes. In such processes, the spatial evolution of a local field is often expected to be smooth. When the physical data on domain boundary is subject to such errors, it is often convenient to model these measurements as a realization of a stochastic process. For instance, the initial condition in Eqn.~\ref{eqn:schrod-IC} can be modeled by a pair of continuous stochastic processes, $U_x, V_x$ where the index representing the spatial coordinate of the PDE domain. The mean and covariance for such processes are given as
\begin{align*}
& \mathbb{E}(U_i) = 2\mathrm{sech}(x = x_i) \\
& \mathbb{E}(V_i) = 0 \\
& \mathrm{Cov}(U_i, U_j) = \mathrm{Cov}(V_i, V_j) = \sigma^2\delta_{ij}
\end{align*}

To obtain a functional estimate of these stochastic processes, Gaussian Process Regression~\cite{gp-book} is a powerful, nonparametric method. Given the set of samples on the initial timeslice, $\mathcal{U}_B$, the DNN structure in Eqn.~\ref{PINN} is replaced by,
\begin{equation}
\tilde{u}(\vec{x}) = \mathbf{NN}_\params\left( \vec{x}; \hat{\mathcal{U}}_B, \mathcal{U}_C,  \mathcal{U}_D \right)
\label{PINN}
\end{equation}
where
\begin{equation}
\hat{\mathcal{U}}_B = \{ (\vec{x}_i^b , \mathcal{B}[\hat{u}(\vec{x}_i^b)] )_{i=1}^{N_b} \}
\end{equation}
and
\begin{equation}
\hat{u}(\vec{x}_i^b) = \mathrm{GP} \left( \vec{x}_i^b | \left\{ (\vec{x}_i^b , u(\vec{x}_i^b) )_{i=1}^{N_b} \right\} \right)
\label{GP-init}
\end{equation}
represents the GP-predicted estimate of the boundary data. The choice of the kernel function, representing the pairwise covariance of observations is given as a sum of RBF and white noise kernels,
\begin{equation}
k(x_i, x_j) = A\exp\left( - \frac{|\vec{x}_i - \vec{x}_j|^2}{2l^2} \right) + \sigma^2\delta_{ij}
\label{eqn-kernel}
\end{equation}
where $A,l,\sigma$ are hyperparameters obtained by maximizing the log-marginal likelihood.

Smoothing techniques are commonly applied in problems where robustness is a desired quality. Compared to other parametric smoothing techniques like fixed order polynominals or smoothing splines, GP regression has often been proved to be more robust against underfitting and overfitting~\cite{gp-fitQ-1,gp-fitQ-2}. Robustness guarantees for GPs have been extensively explored in literature~\cite{robustness-1,robustness-2}. GPs have also been explored in connection with physics-inspired kernel building~\cite{duvenaud2011additive,duvenaud2013structure} and found to be effective in predicting physical phenomena like phase transitions in quantum systems~\cite{vargas2018extrapolating}.

\begin{figure}
\centering
  \subfloat[]{
    \includegraphics[width=0.24\textwidth]{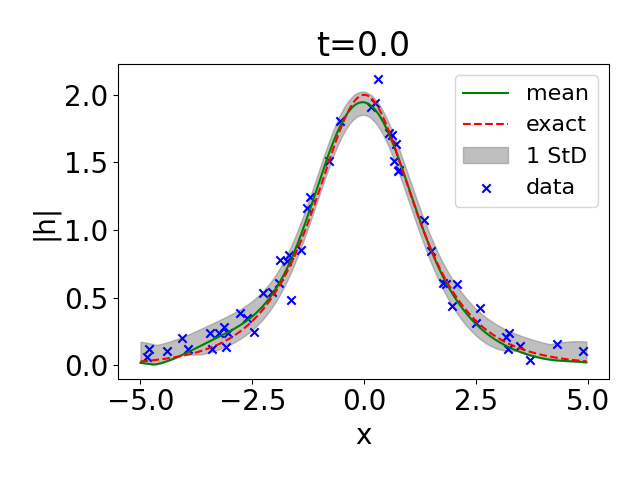}
   \label{H_t_0.0}
  }
  \subfloat[]{
  \includegraphics[width=0.24\textwidth]{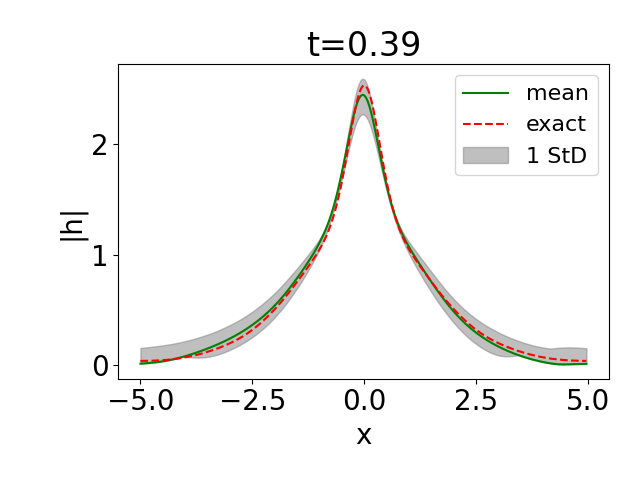}
  \label{H_t_0.39}
  }
  \subfloat[]{
  \includegraphics[width=0.24\textwidth]{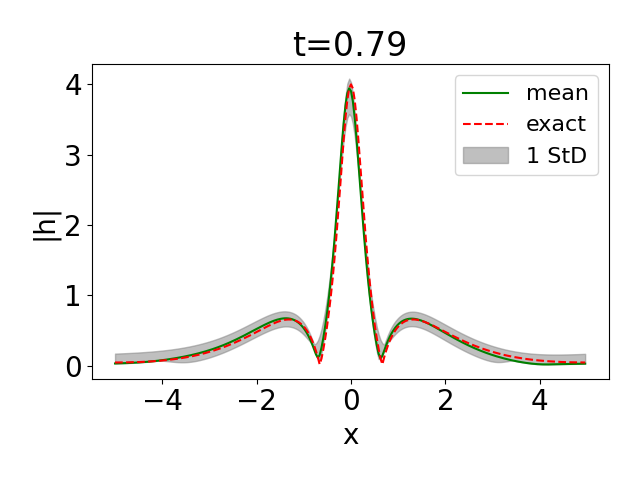}
  \label{H_t_0.78}
  }
  \subfloat[]{
  \includegraphics[width=0.24\textwidth]{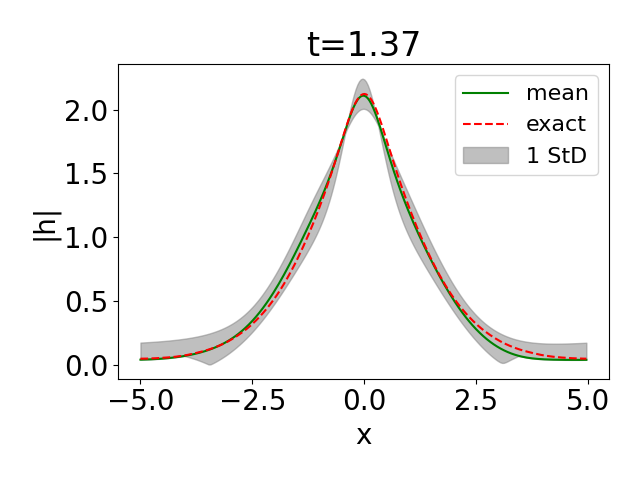}
  \label{H_t_1.37}
  }
  \caption[]{The GP-smoothed PINN-evaluated $|h(x,t)|$ at different timeslices, $t = 0, 0.39, 0.78, 1.37$, from left to right. In this case the training data on the initial timeslice is subject to measurement errors, modeled by a Gaussian random variable with zero mean and a standard deviation of 0.1. GP-based smoothing was used on initial timeslice before training the PINN. The points marked with the blue cross (x) pointer in the leftmost set of plots indicate the samples on the initial timeslice used to fit the GP. The grey band in the subsequent plots represents the uncertainty associated with the PINN-evaluated approximation of  $|h(x,t)|$.}
  \label{fig:PINN-error-GPsmoothed}
\end{figure}

Using a GP-smoothing on the boundary data allows for the PDE solver to regain its performance by training itself over the smoothed data on initial timeslice. Compared to other approaches~\cite{gp-pde-1,gp-pde-2,gp-pde-3} that employ Gaussian Processes to solve differential equations, our method uniquely harnesses the smoothing interpolating functionality of a GP while exploiting the universal approximator feature of a neural network. While Gaussian Processes with a proper choice of a kernel can be very useful in approximating smooth analytical solutions, their $o(n^3)$ complexity makes them infeasible for optimizing such solutions over a large set of collocation points and such complexity grows significantly with high dimensional problems. However, restricting their use on the domain boundary reduces the complexity by an order of magnitude while almost identically recovering the analytical solution. Figure~\ref{fig:PINN-error-GPsmoothed} shows the performance of a GP-smoothed PINN in solving the Schr\"odinger equation, where the DNN can recover the analytical form despite corruption in initial data.

\begin{figure}
\centering
 \subfloat[]{
  \includegraphics[width=0.4\textwidth]{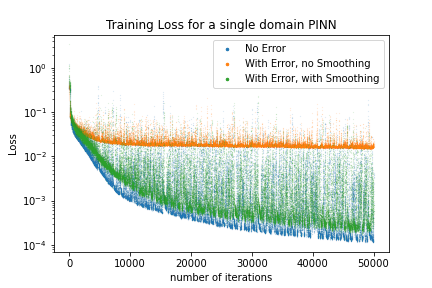}
  \label{70nodes_U_t_0.0}
  }
  \subfloat[]{
  \includegraphics[width=0.4\textwidth]{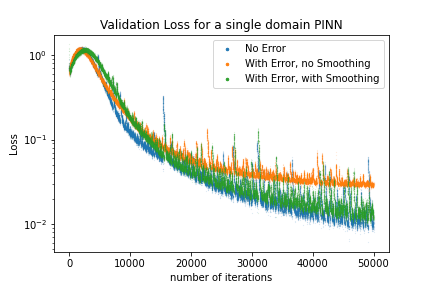}
  \label{100nodes_U_t_0.0}
  }
  \caption[]{The training loss function from Eqn.~\ref{eqn:loss-pinn} (left) and the MSE validation loss from as a function of number of iterations for a PINN solving.}
  \label{Losses-1layer}
\end{figure}

Since the loss function in Eq. \ref{eqn:loss-pinn} is not a direct metric of validating the performance of the the PINN, the validation loss is measured in terms of \textit{mean squared error (MSE)} loss compared with respect to the analytical solution.
In Fig. \ref{Losses-1layer}, we compare the evolution of the loss function during training and the validation MSE loss. It can be seen that GP-smoothed PINN performs almost as well as error-free PINN, and significantly better than a PINN trained with corrupted boundary data but no smoothing.

\subsection{Kernel Selection for Gaussian Processes}

The choice of kernel for fitting a GP to the boundary data is very important- improper choices can lead to underfitting or overfitting and eventually propagate large errors through the PINN architecture. In order to make the best choice for a kernel, we explored a $k$-fold cross-validation technique on the initial time-slice data for the 1D Schr\"odinger Equation. The dataset on initial boundary is split into $k$ equal subsets where $k-1$ of them are used for training and and one subset is kept aside for validation. Training and validation data are used to optimize the GP hyperparameters. We examine the performance of the optimized Gaussian Processes for using the RBF kernel, the Mat\`{e}rn kernel~\cite{gp-book} with $\nu = 0.1, 1.5,$ and $4.0$, and the Rational Quadratic (RQ) kernel~\cite{gp-book}. Each kernel is appended with a localized white noise kernel. The average training and validation MSE losses, measured with respect to a fixed set of noise-corrupted sampling points on the initial timeslice  (Eqn.~\ref{eqn:schrod-IC}), with $k=10$ for different choices of kernels  are summarized in Table~\ref{tab:MSE-kernels}. It can be seen that RBF and RQ kernels have similar performance while the Mat\`{e}rn kernels tend to overfit.

\begin{table}[]
    \centering
    \begin{tabular}{|c|c|c|}
    \hline
        kernel &  MSE loss on training data & MSE loss on validation data\\
        \hline
        RBF & 0.00762 & 0.0110 \\
        Mat\`{e}rn $(\nu = 0.1)$ & $3.2\times 10^{-8} $ & 0.0465 \\
        Mat\`{e}rn $(\nu = 1.5)$ & 0.00598 & 0.0116 \\
        Mat\`{e}rn $(\nu = 4.0)$ & 0.00588 & 0.0126 \\
        Rational Quadratic & 0.00721 & 0.0112 \\
        \hline
    \end{tabular}
    \\ $ $ \\
    \caption{Comparison of average training and validation MSE losses for training and validation data on the domain boundary as given in Eqn.~\ref{eqn:schrod-IC} with $\sigma = 0.1$.}
    \label{tab:MSE-kernels}
\end{table}

\subsection{Evolution of Measurement Uncertainty}

While a generic PINN fails to recover the physics-motivated evolution of noisy boundary data following a PDE, a GP-smoothed PINN not only can recover the physical evolution but also provide a controlled estimate of uncertainty at every point in the spatio-temporal domain. The uncertainty evaluated by a GP-smoothed PINN is obtained by evaluating the deviation in the NN parameters for $\pm 1 \sigma$ variation of the training data on domain boundary

\begin{equation}
\tilde{u}(\vec{x}) \pm \delta\tilde{u}(\vec{x}) = \mathbf{NN}_{\params \pm  \delta\theta}\left( \vec{x}; \hat{\mathcal{U}}_B^\pm, \mathcal{U}_C,  \mathcal{U}_D \right)
\end{equation}
where
\begin{equation}
\hat{\mathcal{U}}_B^\pm = \{ (\vec{x}_i^b , \mathcal{B}[\hat{u}(\vec{x}_i^b) \pm \delta\hat{u}(\vec{x}_i^b)] )_{i=1}^{N_b} \}
\end{equation}

The uncertainty associated with the boundary data, $\delta\hat{u}$ is obtained from the covariance estimate of the optimized GP. The deviation of the NN parameters, $\delta\theta$ can be obtained from minimizing the loss function evaluated with $\hat{\mathcal{U}}_B^\pm$.

\begin{equation}
(\params \pm \delta\theta)^* = \argmin_\params\mathcal{L}_{PINN}(\params; \hat{\mathcal{U}}_B^\pm)
\end{equation}

Analytical estimate of $\delta\theta^*$ is a computationally intractable task since it requires inversion of the very large Hessian matrix $\frac{\partial^2\mathcal{L}}{\partial\theta^2}$. However, a rather inexpensive technique is to start with a PINN architecture with parameters $\theta$ already optimized for the mean value of the boundary data $\hat{\mathcal{U}}_B$ and re-train the network with the modified boundary data. This reoptimization converges more quickly and provides an estimate of evolution of uncertainty at all points of the space-time domain. The evolution of uncertainty for a GP-smoothed PINN evaluated solution of Eqn.~\ref{eqn:pde-schrod} is shown in Figure~\ref{fig:PINN-error-GPsmoothed}. The network was re-trained for an additional 1000 iterations to optimize for the uncertainty bands. In general, the number of additional required to converge for estimating the uncertainty bands depends on the size of the corrupting error, which can be quantitatively estimated from the optimized value of the $\sigma$ parameter in Eqn.~\ref{GP-init}.  

\subsection{Sparse GP (SGP) based Error Correction}

GP-based smoothing can provide robustness for PINNs as shown in the previous section. However, optimizing a GP is an expensive process with a complexity of $o(n^3)$, with $n$ being the number of points considered to optimize the GP. Even though we are restricting the GPs to be optimized only over the domain boundary, this can be still be a major bottleneck for our method for high dimensional problems. As the dimension of domain boundary $\partial{D} \subset \mathbb{R}^{d-1}$ increases, it will require more and more points on the boundary to satisfy the boundary condition.  
Sparse Gaussian Processes have been extensively studied in literature to significantly reduce the complexity for high dimensional problems. 
A multitude of variants of sparsity inducing GPs have found their applications in the context of sample efficient reinforcement learning~\cite{Grande14ReinforcementGP}, deep kriging with big data~\cite{Gadd20DeepGP}, and variational learning of GPs~\cite{tran2016variational}. We consider a hybrid approach for sparsity inducing smoothing GP on the domain boundary following the algorithm suggested in Ref.\cite{InducingPointGP} to obtain inexpensive selection of inducing points (IPs). Originally designed for Sparse Variational GPs (SVGPs), this algorithm is effective in the context of our problem of selecting a smaller subset of IPs on the domain boundary.
\begin{algorithm}
\caption{Selection of IPs for SGP}\label{alg:cap}
\label{alg:SGP}
\begin{algorithmic}
\Procedure{IPSelect}{$n_0,M,X,y,\rho$} 
\State Randomly Select $X_0, y_0$ from $X, y$ with $|X_0| = |y_0| = n_0$ 
\State $k^* = \mathrm{arg max}\log p(y_0 | \mathrm{GP}(k,X_0,y_0))$
\For{$z \in X - X_0$}
\If{$\max{\{k^*(z,x_0)| x_0 \in X_0\}} < \rho$}
\State $X_0 \gets X_0 \cup \{z\}$
\If{$|X_0| = M$}
\State break
\EndIf
\EndIf
\EndFor
\State \textbf{return} $X_0$
\EndProcedure
\end{algorithmic}
\end{algorithm}

\begin{figure}
\centering
\subfloat[]{
    \includegraphics[width=0.24\textwidth]{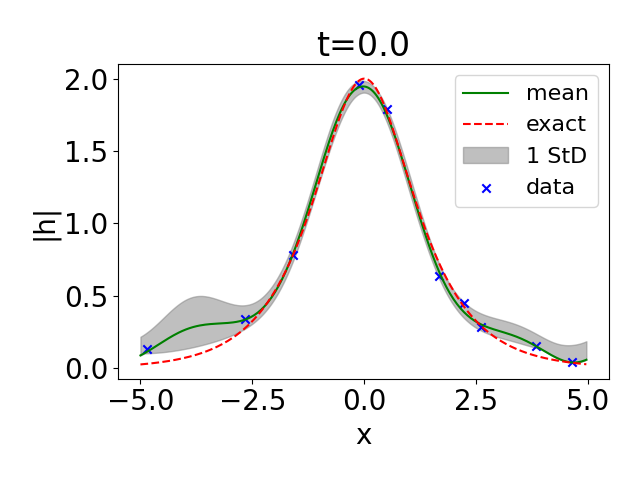}
   \label{10IP_H_t_0.0}
  }
  \subfloat[]{
  \includegraphics[width=0.24\textwidth]{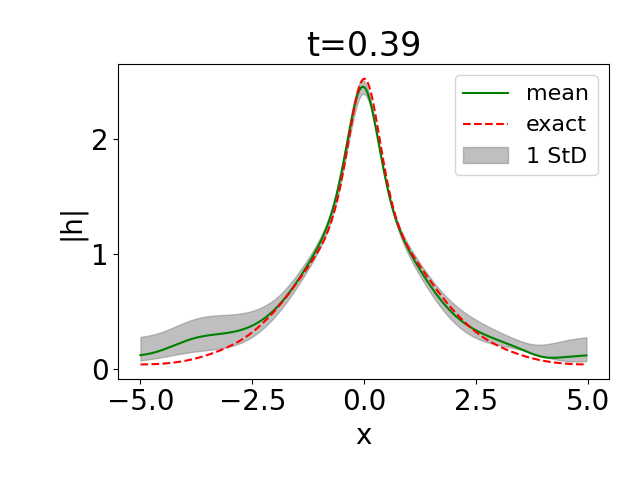}
  \label{10IP_H_t_0.39}
  }
  \subfloat[]{
  \includegraphics[width=0.24\textwidth]{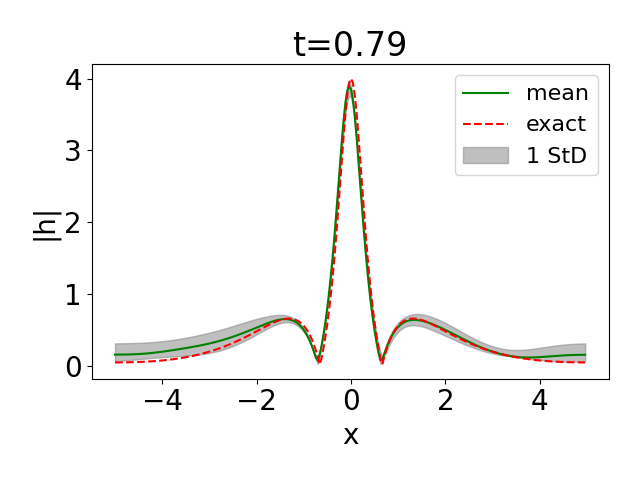}
  \label{10IP_H_t_0.78}
  }
  \subfloat[]{
  \includegraphics[width=0.24\textwidth]{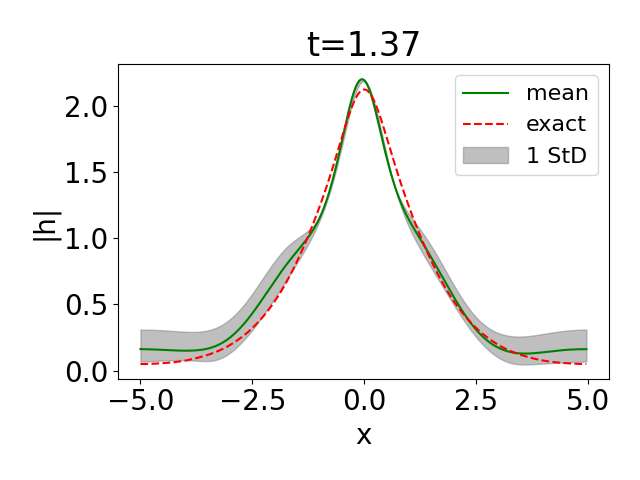}
  \label{10IP_H_t_1.37}
  }
  \\
  \subfloat[]{
    \includegraphics[width=0.24\textwidth]{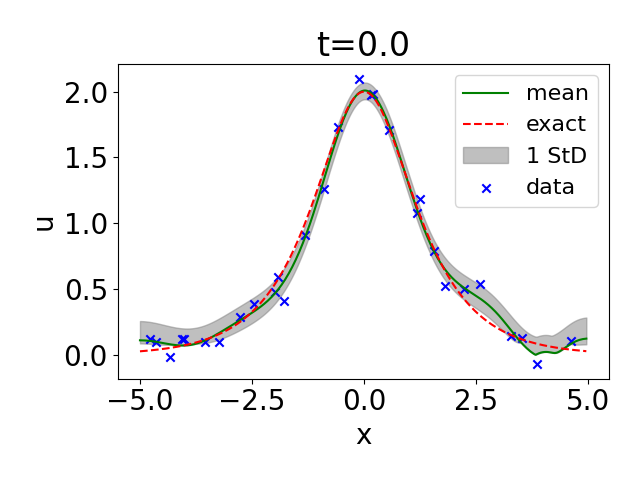}
   \label{30IP_H_t_0.0}
  }
  \subfloat[]{
  \includegraphics[width=0.24\textwidth]{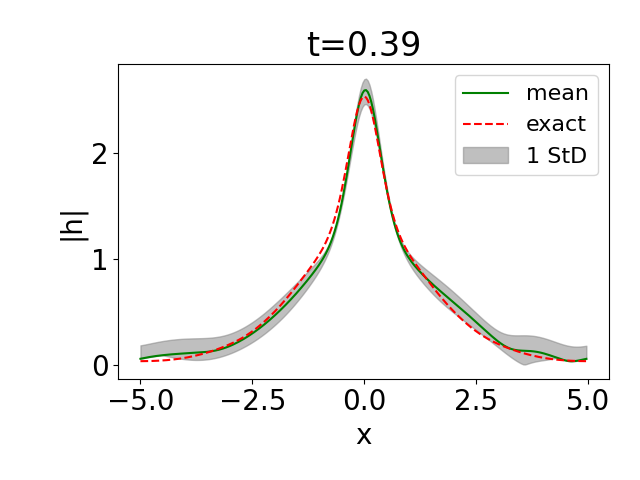}
  \label{30IP_H_t_0.39}
  }
  \subfloat[]{
  \includegraphics[width=0.24\textwidth]{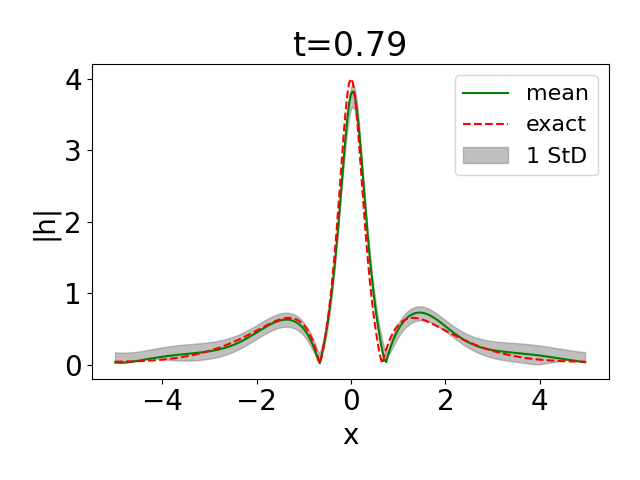}
  \label{30IP_H_t_0.78}
  }
  \subfloat[]{
  \includegraphics[width=0.24\textwidth]{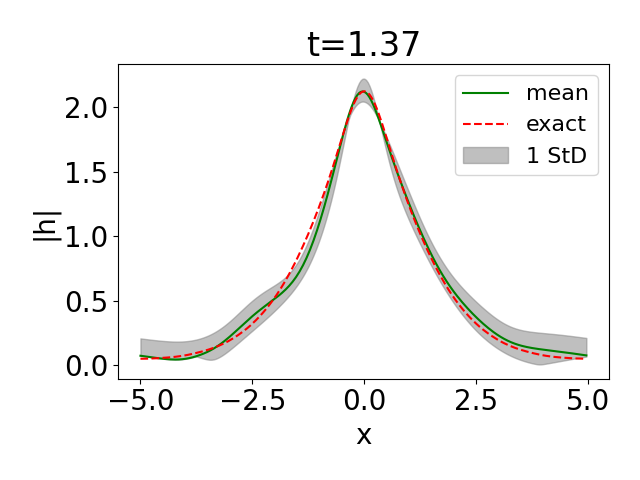}
  \label{30IP_H_t_1.37}
  }
  \caption[]{The SGP-smoothed PINN-evaluated $|h(x,t)|$ at different timeslices, $t = 0, 0.39, 0.79, 1.37$, from left to right. In this case the training data on the initial timeslice is subject to measurement errors, modeled by a Gaussian random variable with zero mean and a standard deviation of 0.1. SGP based smoothing was used on initial timeslice before training the PINN using 10 inducing points for each $u(x,0)$ and $v(x,0)$ for the top row and 29 inducing points for $u(x,0)$ and 20 for $v(x,0)$ for the bottom row. The points marked with the blue cross (x) pointer in the leftmost set of plots indicate the samples on the initial timeslice used to train the PINN. The grey band in the subsequent plots represents the uncertainty associated with the PINN-evaluated approximation of  $|h(x,t)|$.}
  \label{fig:PINN-error-SGPsmoothed}
\end{figure}

The SGP algorithm we use is explained in Algorithm 1. The sparsity optimizations for GP is done in two steps. In the first step, a small number of data points $(n_0)$ are randomly taken to optimize the GP hyperparameters. In the second step, additional IPs are chosen from the data based on the kernel distance between the new IP candidate $(z)$ and the already selected set of IPs $(X_0)$. The new IP candidate is included in $X_0$ if the kernel distance between $z$ and all existing IPs is smaller than some predefined threshold $(\rho)$.
To reduce the complexity of this approach, iterative re-optimization of the kernel hyperparameters is avoided and only after the desired set of IPs have been chosen, the the GP hyperparameters are finally reoptimized to smooth the corrupted dataset on the domain boundary. The total number of IPs chosen is bounded by $M \leq N_{b,t}$. Figure~\ref{fig:PINN-error-SGPsmoothed} shows how sparse GPs can be almost equally useful in recovering the Schr\"odinger field dynamics. When the number of IPs is set too low, e.g. only 10 IPs for both $u(x,0)$ and $v(x,0)$, the recovery of physical dynamics is not as satisfactory. However, with a somewhat larger set of IPs including 29 and 20 IPs for $u(x,0)$ and $v(x,0)$ respectively, the PINN's performance improves significantly and becomes comparable to that of the full GP-smoothed PINN shown in Figure~\ref{fig:PINN-error-GPsmoothed}.

\begin{table}[]
    \centering
    \begin{tabular}{|c|c|}
    \hline
        \textbf{Model} &  \textbf{MSE} \\
        \hline
        PINN (no error) & 0.0105  \\
        PINN ($\sigma = 0.1$) & 0.0289  \\
        PINN ($\sigma = 0.1$, $L_1$ regularization with $\lambda = 10^{-4}$) & 0.1613  \\
        PINN ($\sigma = 0.1$, $L_2$ regularization with $\lambda = 10^{-4}$) & 0.2681  \\
        cPINN-2 (no error) & 0.2745  \\
        cPINN-2 ($\sigma = 0.1$, no smoothing) & 0.4782  \\
        cPINN-3 (no error) & 0.0258  \\
        cPINN-3 ($\sigma = 0.1$, no smoothing) & 0.4178  \\
        GP-smoothed PINN ($\sigma = 0.1$, 50 IPs for $u$ and $v$) & 0.0125 \\
        SGP-smoothed PINN ($\sigma = 0.1$, 10 IPs for $u$ and $v$) & 0.0231 \\
        SGP-smoothed PINN ($\sigma = 0.1$, 29 and 20 IPs for $u$ and $v$) & 0.0123 \\
        \hline
    \end{tabular}
    \\ $ $ \\
    \caption{Comparison of PINNs using different strategies for robustness to solve the 1D nonlinear Schr\"odinger equation. The introduction of error in the initial condition causes a significant increase in MSE for the standard PINN. GP-smoothing reduces the MSE to nearly as low as the PINN with no error. SGP-smoothing is also effective in reducing error and uses fewer inducing points (IPs). However, if the SGP does not have a sufficient number of IPs the error increases as seen when 10 IPs are used. Multiple domain cPINNs have worse performance. Results quoted for $L_1$ and $L_2$ regularizations are taken from the best performance observed over choices of $\lambda \in \{ 10^{-n}\}_{n=1}^5$.
    }
    \label{tab:MSE-schrod}
\end{table}

Table~\ref{tab:MSE-schrod} summarizes the validation MSE obtained with different models and compare them with the benchmark model of a vanilla PINN with no errors. While the performance of a PINN significantly deteriorates with the introduction of even modest errors with $\sigma = 0.1$, both GP-smoothed PINN and SVGP-smoothed PINN perform similar to the benchmark model.

\begin{figure}[!h]
    \centering
    \subfloat[]{
        \includegraphics[width=0.24\textwidth]{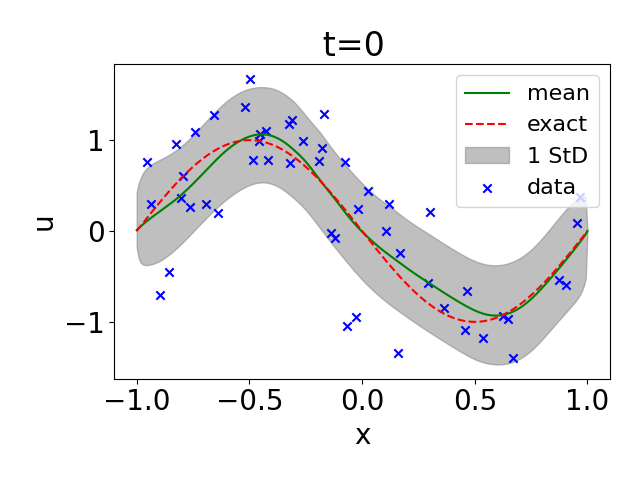}
    }
    \subfloat[]{
        \includegraphics[width=0.24\textwidth]{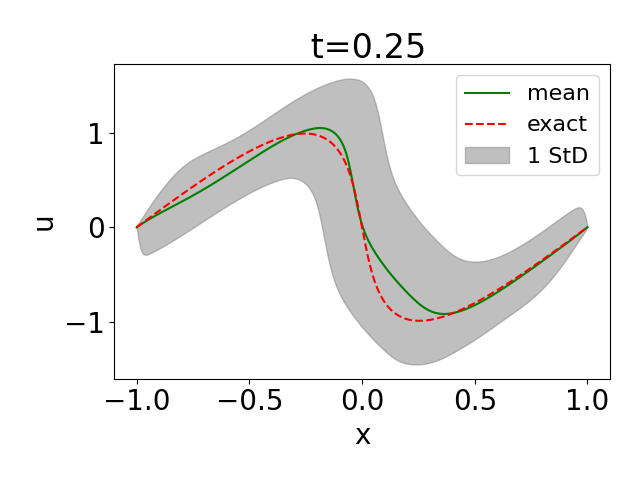}
    }
    \subfloat[]{
        \includegraphics[width=0.24\textwidth]{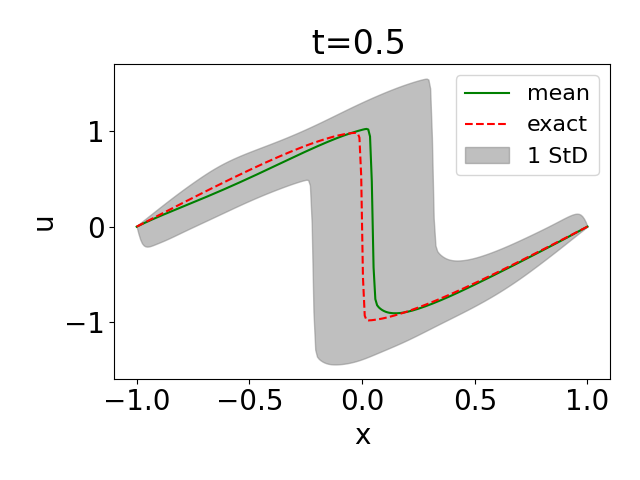}
    }
    \subfloat[]{
        \includegraphics[width=0.24\textwidth]{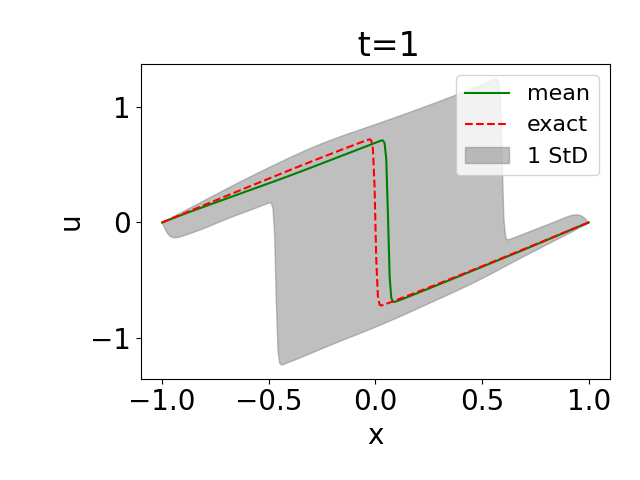}
    }
    \\
    \subfloat[]{
        \includegraphics[width=0.24\textwidth]{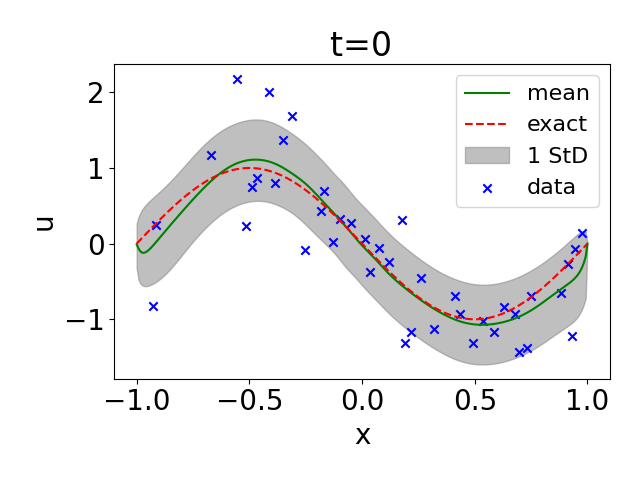}
    }
    \subfloat[]{
        \includegraphics[width=0.24\textwidth]{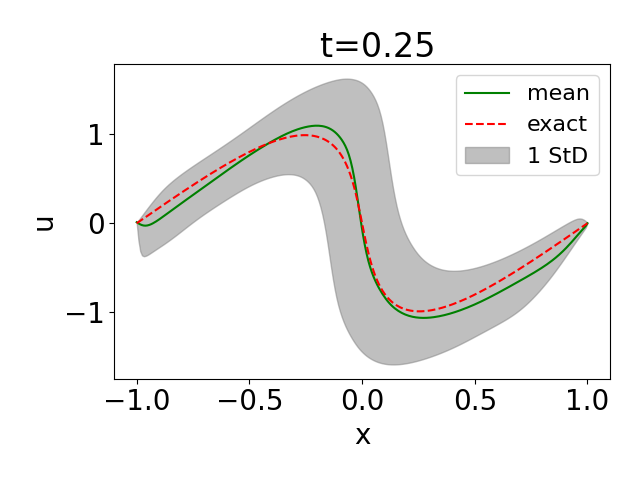}
    }
    \subfloat[]{
        \includegraphics[width=0.24\textwidth]{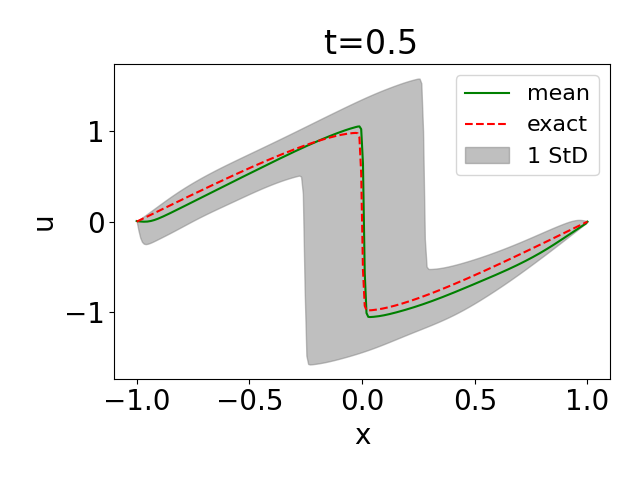}
    }
    \subfloat[]{
        \includegraphics[width=0.24\textwidth]{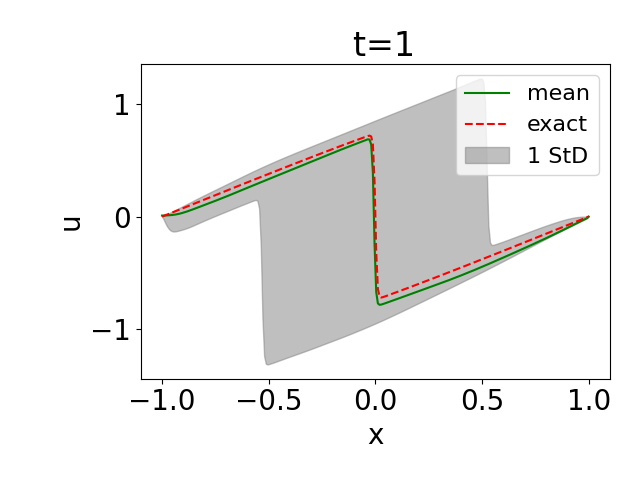}
    }
    \\
    \subfloat[]{
        \includegraphics[width=0.24\textwidth]{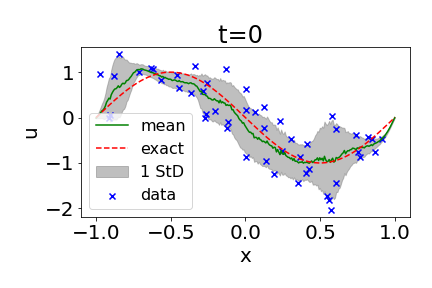}
    }
    \subfloat[]{
        \includegraphics[width=0.24\textwidth]{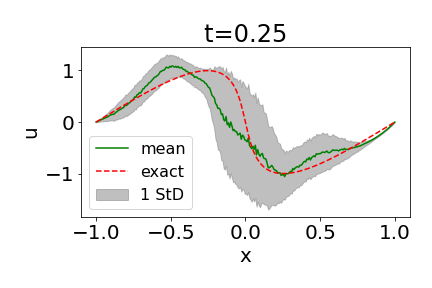}
    }
    \subfloat[]{
        \includegraphics[width=0.24\textwidth]{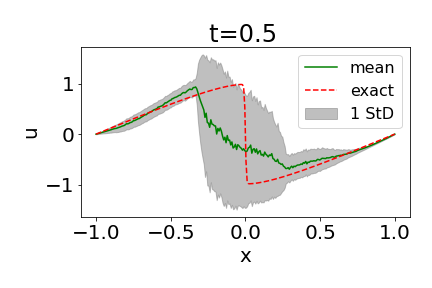}
    }
    \subfloat[]{
        \includegraphics[width=0.24\textwidth]{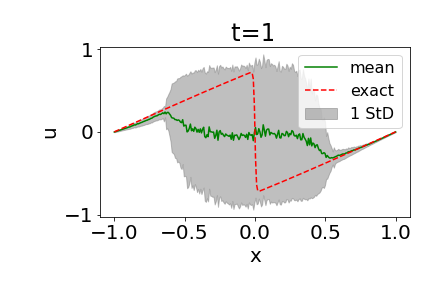}
    }
    \caption{The PINN-evaluated solution to Burgers' equation when measurements on initial timeslice is corrupted with zero-mean, $\sigma=0.5$ Gaussian noise for GP-smoothed PINN (top), SGP-smoothed PINN (middle), and  UQ-PINN~\cite{pinn-adversarial} (bottom), at time-slices $t=0,0.25,0.51,1$, from left to right. For the GP- and SGP-smoothed PINNs uncertainty bounds are calculated by retraining the PINN using the initial condition of the GP/SGP mean function plus or minus one standard deviation.
    }
    \label{fig:burger-0.5}
\end{figure}

We demonstrate the effectiveness of GP and SGP smoothing in recovering the physical field dynamics for 1D Burgers' equation in Figures~\ref{fig:burger-0.5}(a)-(d) and Figures~\ref{fig:burger-0.5}(e)-(h). The SGP employed 41 IPs on the initial timeslice and shows remarkable performance recovery. We also compare the results from the UQ-PINN architecture proposed in Ref.~\cite{pinn-adversarial} and both GP and SGP smoothed PINNs perform noticeably better than the solution obtained from the UQ-PINN architecture. Table~\ref{tab:MSE-burger} summarizes the validation MSE loss obtained from different PINN architectures and it can be seen that both GP and SGP smoothed recover a similar level of accuracy as observed by the error-free PINN.     

\begin{table}[]
    \centering
    \begin{tabular}{|c|c|}
     \hline
        \textbf{Model} &  \textbf{MSE}  \\
        \hline
        PINN (no error) & 0.0116  \\
        PINN ($\sigma = 0.5$) & 0.1982  \\
        PINN ($\sigma = 0.1$, $L_1$ regularization with $\lambda = 10^{-4}$) & 0.0392  \\
        PINN ($\sigma = 0.1$, $L_2$ regularization with $\lambda = 10^{-4}$) & 0.0293  \\
        PINN ($\sigma = 0.5$, Cole-Hopf regularizer) & 0.1125  \\
        cPINN-2 (no error) & 0.0161  \\
        cPINN-2 ($\sigma = 0.5$, no smoothing) & 0.0834   \\
        cPINN-2 ($\sigma = 0.5$, Cole-Hopf regularizer) & 0.0891  \\
        cPINN-3 (no error) & 2.782e-5   \\
        cPINN-3 ($\sigma = 0.5$, no smoothing) & 0.0854   \\
        cPINN-3 ($\sigma = 0.5$, Cole-Hopf regularizer) & 0.0329  \\
        UQ-PINN~\cite{pinn-adversarial} ($\sigma = 0.5)$ & 0.1248  \\
        GP-smoothed PINN ($\sigma = 0.5$, 50 IPs) & 0.0384  \\
        SGP-smoothed PINN ($\sigma = 0.5$, 41 IPs) & 0.0080   \\
    \hline    
    \end{tabular}
    \\ $ $ \\
    \caption{Comparison of PINNs using different strategies for robustness to solve the 1D Burgers' equation. The introduction of error in the initial condition causes a significant increase in MSE for the standard PINN. GP-smoothing reduces the MSE to nearly as low as the PINN with no error. SGP-smoothing is also effective in reducing error and uses fewer inducing points (IPs). Results quoted for $L_1$ and $L_2$ regularizations are taken from the best performance observed over choices of $\lambda \in \{ 10^{-n}\}_{n=1}^5$.
    }
    \label{tab:MSE-burger}
\end{table}

%% file: tex-paper/5AddnlEx.tex
\section{Additional Examples}

To demonstrate the effectiveness of GP and SGP smoothing for higher dimensional PDEs, we consider a couple of 2D PDEs in this section.

\subsection{2D Heat Equation}

The 2D heat equation and the corresponding spatio-temporal boundary conditions are given as:
\begin{align}
    & \partialD{u}{t} = \partialD{^2 u}{x^2} + \partialD{^2 u}{y^2} \label{eqn:pde-heat2d} \\
    & u(x,y,0) = 3\sin(\pi x)\sin(\pi y) + \sin(3\pi x)\sin(\pi y) \label{eqn:heat2dIC} + \Theta_u\epsilon^u \\
    & u(0,y,t) = u(1,y,t) = u(x,0,t) = u(x,1,t) = 0 \label{eqn:heat2dBC}
\end{align}
where the domain boundary is given as $(x,y,t) \in [0, 1] \times [0,1] \times [0, 0.1]$ and $\Theta_u$ is the acceptance function for the noise term in the initial condition. The analytical solution to this equation is given as $u(x,y,t) = 3\sin(\pi x)\sin(\pi y)e^{-2\pi^2t^2} + \sin(3\pi x)\sin(\pi y)e^{-10\pi^2t^2}$. An MLP with four hidden layers, each with 256 nodes, has been used. The physics is enforced with $N_c = 50000$ collocation points. 64 points are chosen on each of the four spatial boundaries and 1024 points on the initial timeslice for the initial condition, giving a total of  $N_b = 1280$ points on the spatio-temporal boundary. Like the previous examples, the loss function is constructed according to Eq.~\ref{eqn:loss-pinn} with $\alpha_{()} = 1.0$. For the SGP process, the IPs are chosen from the pool of 1024 points on the initial time slice according to Algorithm 1 with the number of IPs bounded by $M = 768$. The models are trained for 20000 epochs with \textsc{Adam} optimizer with a learning rate of $10^{-3}$.

\begin{figure}[!h]
  \begin{minipage}[c]{0.58\columnwidth}
    \centering
    \includegraphics[width=\columnwidth]{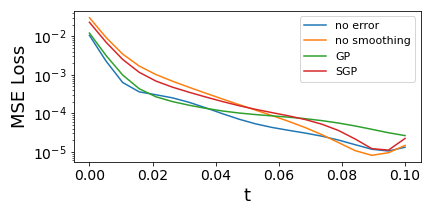}
  \end{minipage}%
  \begin{minipage}[c]{0.150\columnwidth}
    \centering
\begin{tabular}[b]{ |c|c| }
  \hline
  Model & MSE \\
  \hline
  PINN (no error) & 7.6e-4 \\
  PINN ($\sigma = 1.0$) & 2.4e-3 \\
  GP-PINN ($\sigma = 1.0$, 1024 IPs) & 9.2e-4 \\
  SGP-PINN ($\sigma = 1.0$, 602 IPs) & 1.8e-3\\
  \hline
\end{tabular}
\end{minipage}
\caption{The time evolution of the MSE loss for different models used in solving the 2D heat equation. Except the vanilla PINN (no error), all models were trained with data sampled from the initial timeslice corrupted with additive Gaussian noise with zero mean and $\sigma = 1.0$. MSE error evaluated over 50k points chosen over the entire spatio-temporal domain for the different models is given in the accompanying table.}
\label{fig:heat2d-MSE-results}
\end{figure}

\begin{figure}
\centering
  \subfloat[]{
  \includegraphics[width=0.24\textwidth]{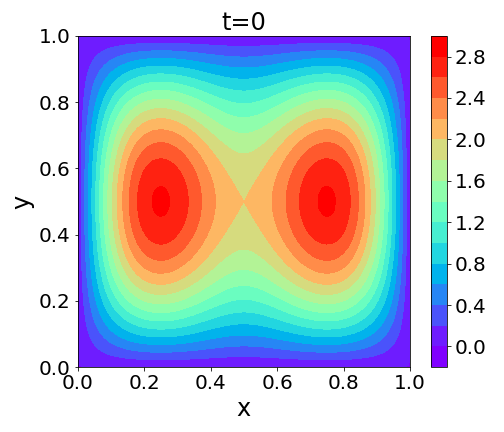}
  \label{fig:heat2d_ICexact}
  }
  \subfloat[]{
  \includegraphics[width=0.24\textwidth]{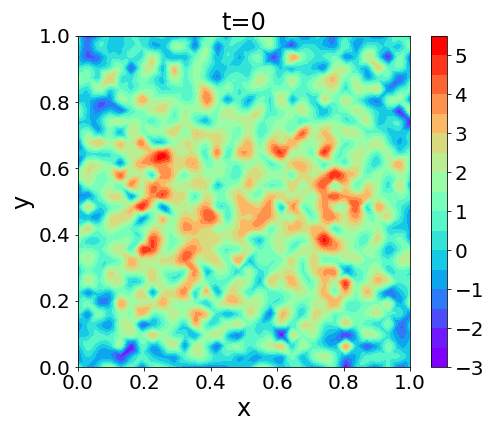}
  \label{fig:heat2d_ICnosmoothing}
  }
  \subfloat[]{
  \includegraphics[width=0.24\textwidth]{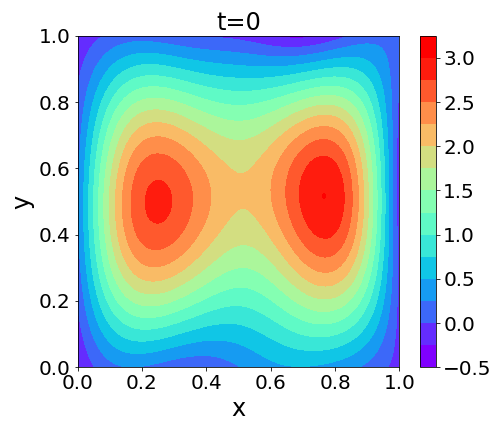}
  \label{fig:heat2d_ICgp}
  }
  \subfloat[]{
  \includegraphics[width=0.24\textwidth]{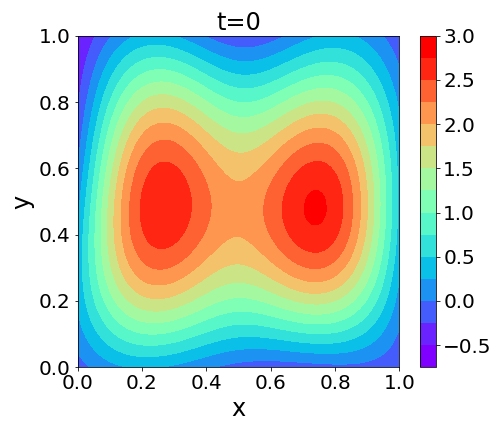}
  \label{fig:heat2d_ICsgp}
  } \\
  \subfloat[]{
  \includegraphics[width=0.24\textwidth]{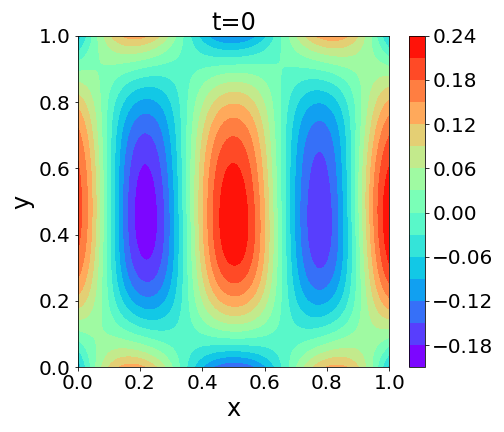}
  \label{fig:heat2d_exact_t0_noise}
  }
  \subfloat[]{
  \includegraphics[width=0.24\textwidth]{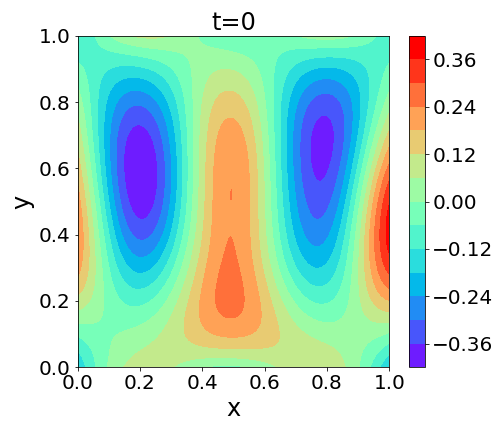}
  \label{fig:heat2d_nosmoothing_t0_noise}
  }
  \subfloat[]{
  \includegraphics[width=0.24\textwidth]{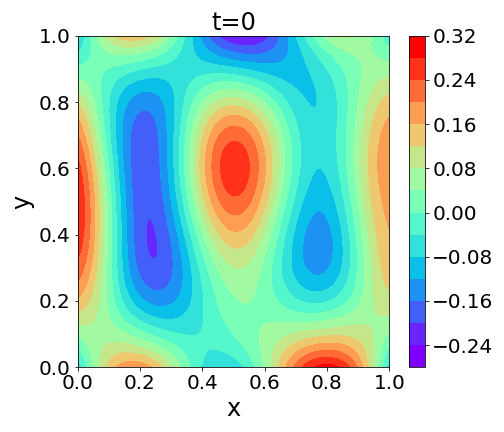}
  \label{fig:heat2d_gp_t0_noise}
  }
  \subfloat[]{
  \includegraphics[width=0.24\textwidth]{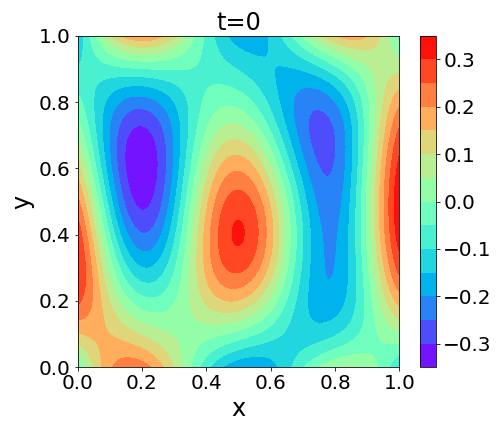}
  \label{fig:heat2d_sgp_t0_noise}
  }
  \caption[]{The initial condition at $t=0$ for 2D heat equation for \protect\subref{fig:heat2d_ICexact} error-free PINN, \protect\subref{fig:heat2d_ICnosmoothing} noisy PINN without smoothing,
  \protect\subref{fig:heat2d_ICgp} noisy PINN with GP smoothing,
  \protect\subref{fig:heat2d_ICsgp} noisy PINN with SGP smoothing. Except the vanilla PINN (no error), all models were trained with data sampled from the initial timeslice corrupted with additive Gaussian noise with zero mean and $\sigma = 1.0$. The figures in the bottom row show the error in PINN-evaluated solution at the initial timeslice for the corresponding architecture in the top row. }
  \label{fig:heat2d_IC}
\end{figure}

\begin{figure}
\centering
  \subfloat[]{
  \includegraphics[width=0.2\textwidth]{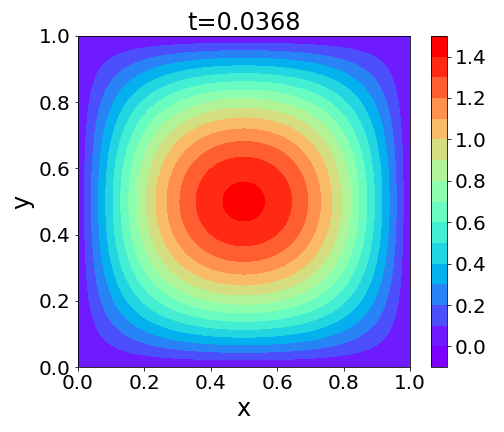}
  \label{fig:heat2d_t=0.0368exact}
  }
  \subfloat[]{
  \includegraphics[width=0.2\textwidth]{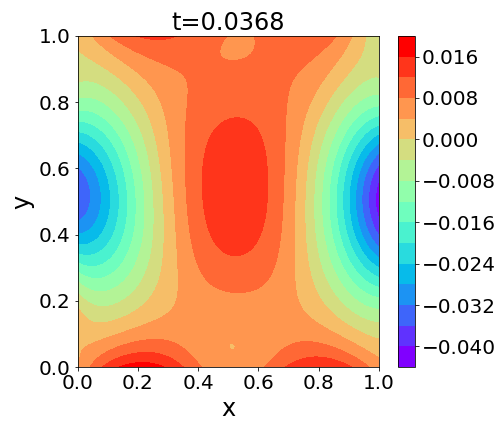}
  \label{fig:heat2d_t=0.0368exact_noise}
  }
  \subfloat[]{
  \includegraphics[width=0.2\textwidth]{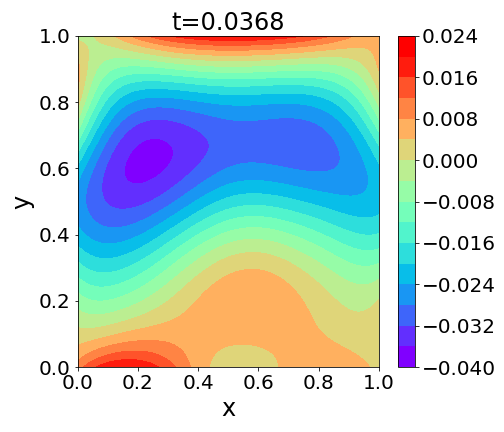}
  \label{fig:heat2d_t=0.0368nosmoothing_noise}
  }
  \subfloat[]{
  \includegraphics[width=0.2\textwidth]{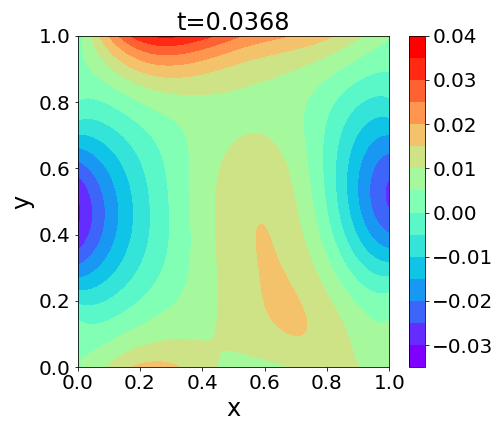}
  \label{fig:heat2d_t=0.0368gp_noise}
  }
  \subfloat[]{
  \includegraphics[width=0.2\textwidth]{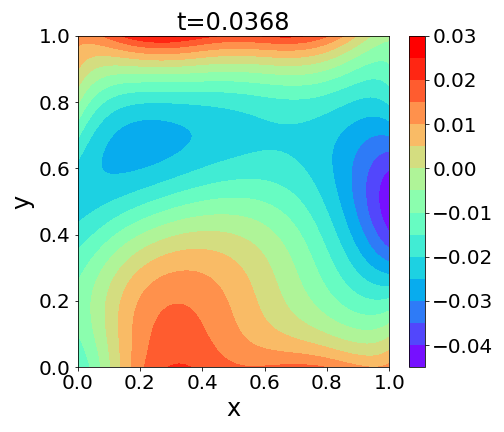}
  \label{fig:heat2d_t=0.0368sgp_noise}
  } \\
  \subfloat[]{
  \includegraphics[width=0.2\textwidth]{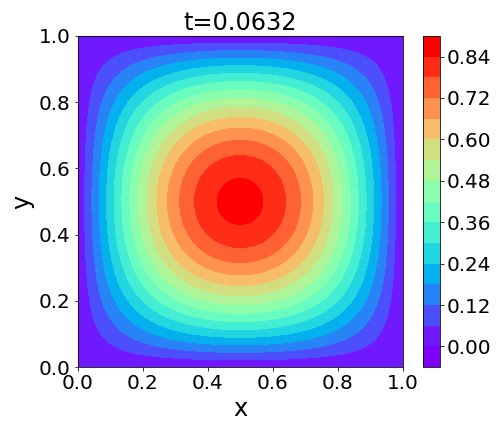}
  \label{fig:heat2d_t=0.0632exact}
  }
  \subfloat[]{
  \includegraphics[width=0.2\textwidth]{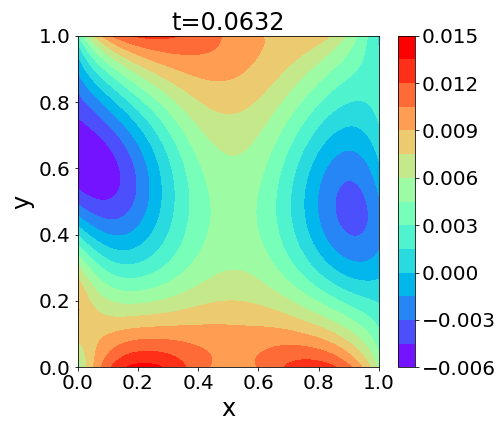}
  \label{fig:heat2d_t=0.0632exact_noise}
  }
  \subfloat[]{
  \includegraphics[width=0.2\textwidth]{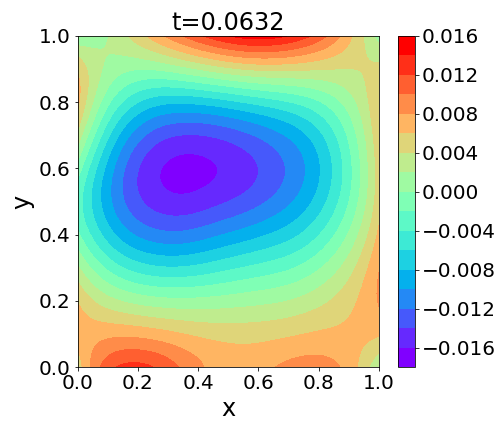}
  \label{fig:heat2d_t=0.0632nosmoothing_noise}
  }
  \subfloat[]{
  \includegraphics[width=0.2\textwidth]{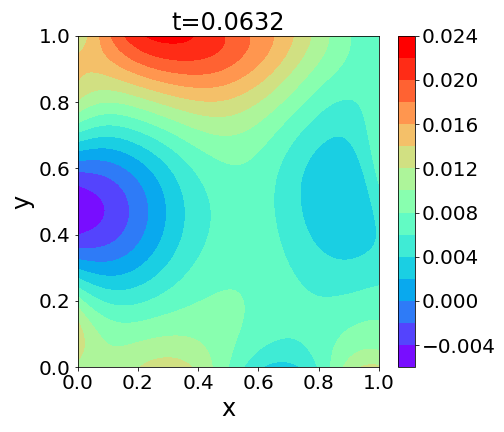}
  \label{fig:heat2d_t=0.0632gp_noise}
  }
  \subfloat[]{
  \includegraphics[width=0.2\textwidth]{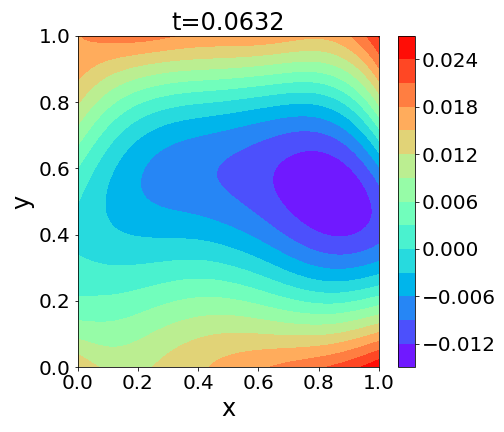}
  \label{fig:heat2d_t=0.0632sgp_noise}
  }
  
  \caption[]{The leftmost column shows the exact solution for 2D heat equation at $t = 0.0368$ (top row) and $t= 0.0632$ (bottom row). The remaining columns show the point-wise error in PINN-evaluated solution for noise-free PINN (second column), noisy PINN without smoothing (third column), GP-PINN (fourth column), and SGP-PINN (final i.e. fifth column). Except the vanilla PINN (no error), all models were trained with data sampled from the initial timeslice corrupted with additive Gaussian noise with zero mean and $\sigma = 1.0$. }
  \label{fig:heat2d_timeslices}
\end{figure}

As shown in Figure~\ref{fig:heat2d-MSE-results}, GP-smoothing recovers the performance of the error-free PINN and SGP also considerably brings down the MSE when compared to that of the PINN trained with noisy data without any smoothing applied. We can see the initial condition each PINN architecture is trained with along with the point-wise error estimate in the PINN's solution for different models. The smoothing effect on the initial timeslice can be seen in Figure~\ref{fig:heat2d_IC}, where we can see that while the noisy initial condition almost completely obliterates the distributive feature of $u(x,t=0)$, smoothing with GP or SGP allows its significant recovery. This translates into better convergence to actual solution for the latter couple of models on both initial and latter timeslices (Figure~\ref{fig:heat2d_timeslices}). 

\subsection{2D Burgers' Equation}
The two dimensional Burgers' equation is given by the following pair of PDEs-
\begin{align}
    \partialD{u}{t} + u\partialD{u}{x} + v\partialD{u}{y} &= \nu \left(\partialD{^2 u}{x^2} + \partialD{^2 u}{y^2} \right) \nonumber \\
    \partialD{v}{t} + u\partialD{v}{x} + v\partialD{v}{y} &= \nu \left(\partialD{^2 v}{x^2} + \partialD{^2 v}{y^2} \right) 
    \label{eqn:burgers-2d}
\end{align}
where we consider $\nu = \frac{0.01}{\pi}$ and train the network to learn the following analytical solution~\cite{fletcher1983generating, zhu2010numerical}-
\begin{align}
    u(x,y,t) &= \frac{3}{4} - \frac{1}{4\left( 1 + \exp{\left(\frac{-t-4x+4y}{32\nu}\right)} \right)} \label{eqn:burgers-2d-exact-u} \\
    v(x,y,t) &= \frac{3}{4} + \frac{1}{4\left( 1 + \exp{\left(\frac{-t-4x+4y}{32\nu}\right)} \right)} \label{eqn:burgers-2d-exact-v}
\end{align}

\begin{figure}[!h]
  \begin{minipage}[c]{0.58\columnwidth}
    \centering
    \includegraphics[width=\columnwidth]{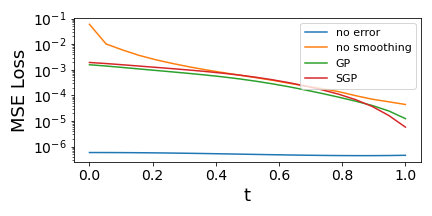}
  \end{minipage}%
  \begin{minipage}[c]{0.150\columnwidth}
    \centering
\begin{tabular}[b]{ |c|c| }
  \hline
  Model & MSE \\
  \hline
  PINN (no error) & 5.16e-7 \\
  PINN ($\sigma = 0.5$) & 4.4e-2 \\
  GP-PINN ($\sigma = 0.5$, 1024 IPs) & 5.3e-4 \\
  SGP-PINN ($\sigma = 0.5$, 512 IPs) & 7.0e-4\\
  \hline
\end{tabular}
\end{minipage}
\caption{The figure demonstrates the time evolution of the MSE loss for different models used in solving the 2D Burgers' Equation. Except the vanilla PINN (no error), all models were trained with data sampled from the initial timeslice corrupted with additive Gaussian noise with zero mean and $\sigma = 0.5$. MSE error evaluated over 50k points chosen over the entire spatio-temporal domain for the different models is given in the accompanying table}
\label{fig:burgers2d-MSE-results}
\end{figure}

The domain boundary is chosen as $(x,y,t) \in [0, 1] \times [0,1] \times [0, 1]$. The network is trained with the initial condition sampled from the functions $u(x,y,0) + \Theta_u\epsilon^u$ and $v(x,y,0) + \Theta_v\epsilon^v$ respectively for $u$ and $v$ where  $\Theta_u$ and $\Theta_v$ are the acceptance functions for the noise terms in the initial condition. The spatial boundary conditions are obtained from plugging in the boundary coordinates in the analytical solution given in Equations~\ref{eqn:burgers-2d-exact-u}~and~\ref{eqn:burgers-2d-exact-v} . An MLP with four hidden layers, each with 256 nodes, has been used to simultaneously predict the two fields. The physics is enforced with $N_c = 50000$ collocation points. We choose 64 points  on each of the four spatial boundaries and 1024 points on the initial timeslice to enforce the spatio-temporal boundary condition with a total $N_b = 1280$ measurements.  The choice of loss function and optimizer follows the example of the previous examples. A pool of 1024 uniformly sampled points on the initial timeslice are used for SGP and IPs are chosen according to Algorithm 1 with the number of IPs bounded by $M = 768$. The resulting performances of the four models after training for 20000 epochs are shown in Figure~\ref{fig:burgers2d-MSE-results}. As we can see from the accompanying table in Figure~\ref{fig:burgers2d-MSE-results}, the noise-free PINN replicates the analytical solution almost perfectly and when error is introduced, similar level of performance cannot be retrieved. However, the MSE is noticably reduced by the smoothing performed by GP and SGP.

\begin{figure}
\centering
  \subfloat[]{
  \includegraphics[width=0.24\textwidth]{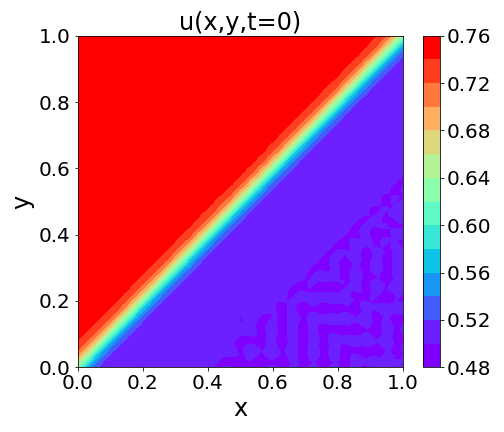}
  \label{fig:burgers2d_u_ICexact}
  }
  \subfloat[]{
  \includegraphics[width=0.24\textwidth]{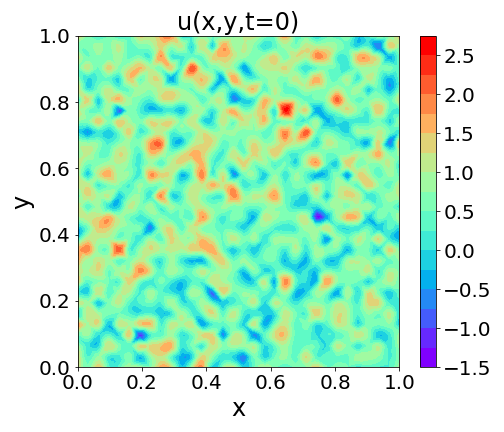} 
  \label{fig:burgers2d_u_ICnosmoothing}
  }
  \subfloat[]{
  \includegraphics[width=0.24\textwidth]{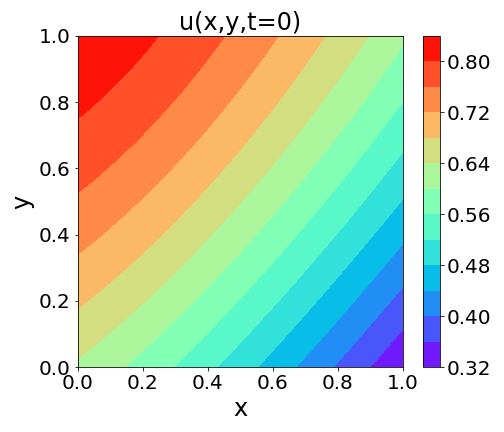}
  \label{fig:burgers2d_u_ICgp}
  }
  \subfloat[]{
  \includegraphics[width=0.24\textwidth]{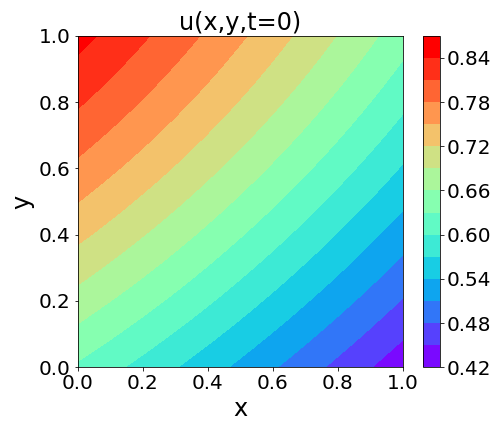}
  \label{fig:burgers2d_u_ICsgp}
  } \\
  \subfloat[]{
  \includegraphics[width=0.24\textwidth]{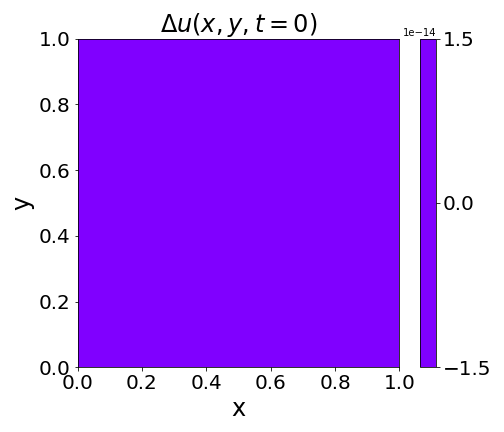}
  \label{fig:burgers2d_exact_t0_u_noise}
  }
  \subfloat[]{
  \includegraphics[width=0.24\textwidth]{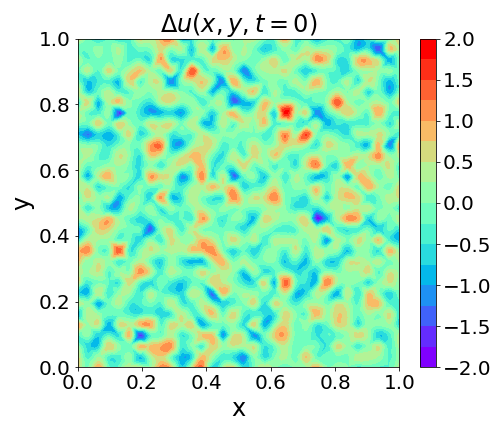}
  \label{fig:burgers2d_nosmoothing_t0_u_noise}
  }
  \subfloat[]{
  \includegraphics[width=0.24\textwidth]{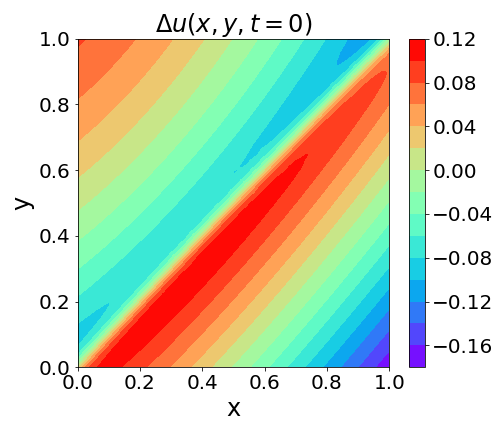}
  \label{fig:burgers2d_gp_t0_u_noise}
  }
  \subfloat[]{
  \includegraphics[width=0.24\textwidth]{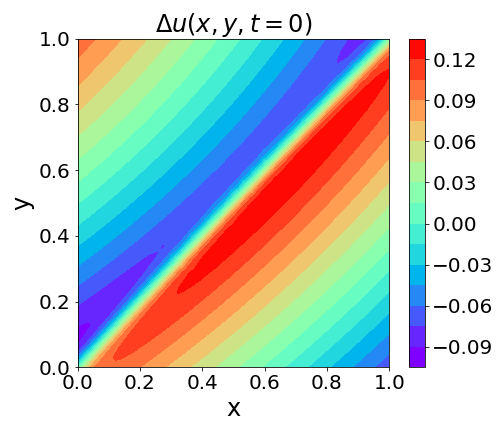}
  \label{fig:burgers2d_sgp_t0_u_noise}
  } \\
  \subfloat[]{
  \includegraphics[width=0.24\textwidth]{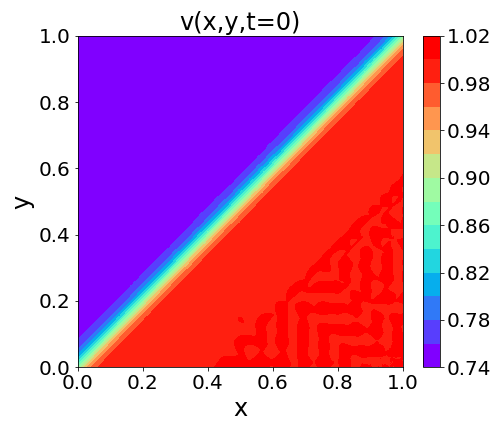}
  \label{fig:burgers2d_v_ICexact}
  }
  \subfloat[]{
  \includegraphics[width=0.24\textwidth]{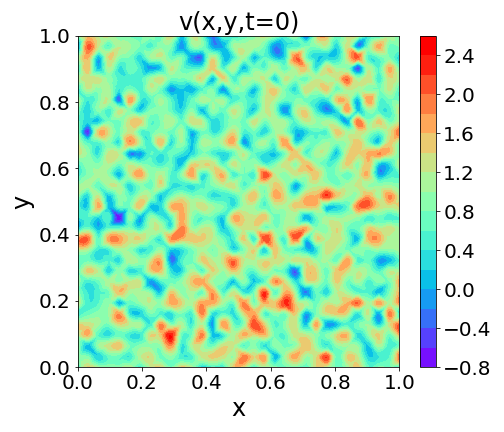} 
  \label{fig:burgers2d_v_ICnosmoothing}
  }
  \subfloat[]{
  \includegraphics[width=0.24\textwidth]{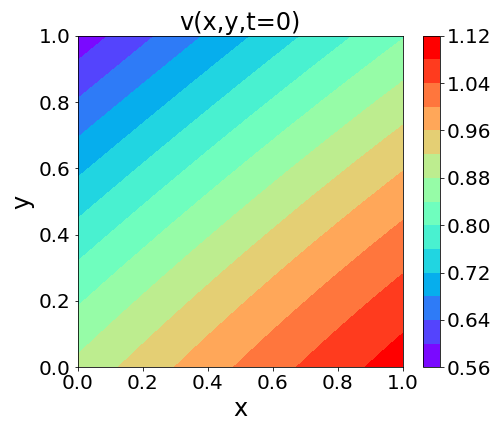}
  \label{fig:burgers2d_v_ICgp}
  }
  \subfloat[]{
  \includegraphics[width=0.24\textwidth]{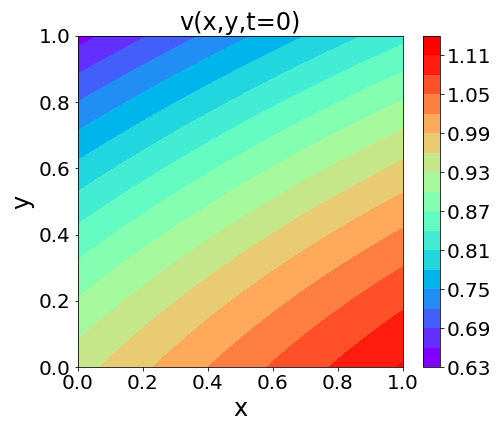}
  \label{fig:burgers2d_v_ICsgp}
  } \\
  \subfloat[]{
  \includegraphics[width=0.24\textwidth]{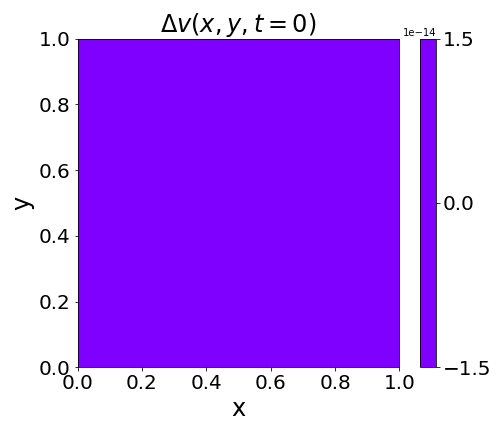}
  \label{fig:burgers2d_exact_t0_v_noise}
  }
  \subfloat[]{
  \includegraphics[width=0.24\textwidth]{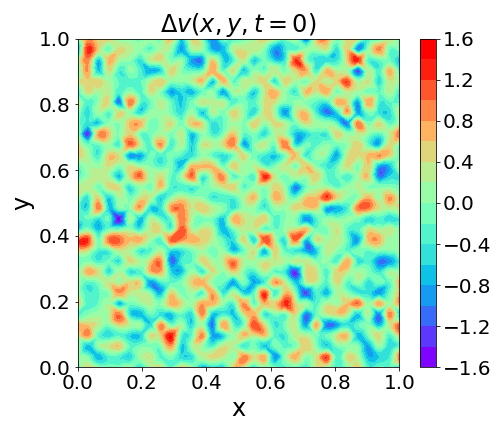}
  \label{fig:burgers2d_nosmoothing_t0_v_noise}
  }
  \subfloat[]{
  \includegraphics[width=0.24\textwidth]{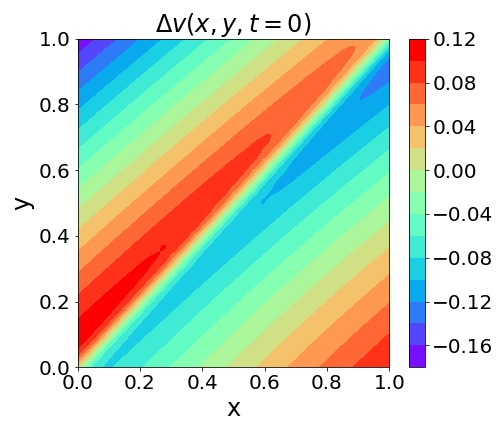}
  \label{fig:burgers2d_gp_t0_v_noise}
  }
  \subfloat[]{
  \includegraphics[width=0.24\textwidth]{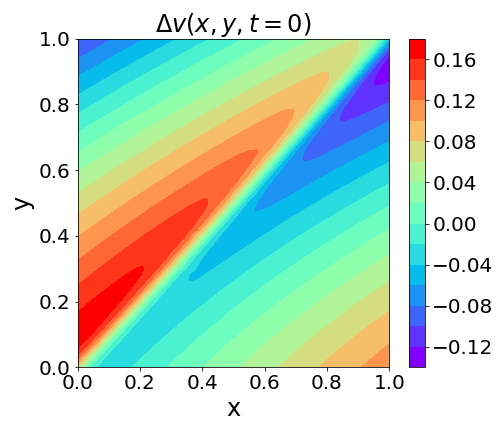}
  \label{fig:burgers2d_sgp_t0_v_noise}
  }
  \caption[]{The initial condition $u(x,y,t=0)$  for 2D Burgers' equation for \protect\subref{fig:burgers2d_u_ICexact} error-free PINN, \protect\subref{fig:burgers2d_u_ICnosmoothing} noisy PINN without smoothing,
  \protect\subref{fig:burgers2d_u_ICgp} noisy PINN with GP smoothing,
  \protect\subref{fig:burgers2d_u_ICsgp} noisy PINN with SGP smoothing. The figures in the second row show the error in PINN-evaluated solution at the initial timeslice for the corresponding architecture in the top row. The final two rows show the equivalent distributions for the $v(x,y,t)$ field. Except the vanilla PINN (no error), all models were trained with data sampled from the initial timeslice corrupted with additive Gaussian noise with zero mean and $\sigma = 0.5$.}
  \label{fig:burgers2d_IC}
\end{figure}

\begin{figure}
\centering
  \subfloat[]{
  \includegraphics[width=0.2\textwidth]{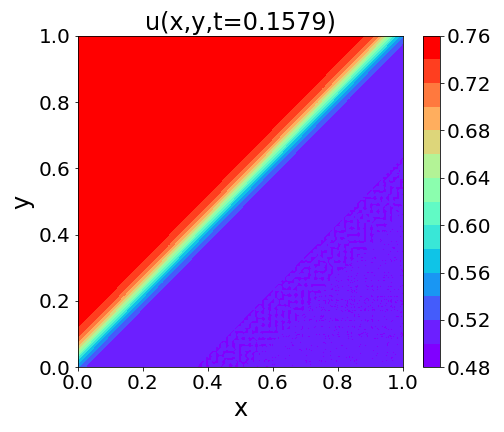}
  \label{fig:burgers2d_t=0.1579exact}
  }
  \subfloat[]{
  \includegraphics[width=0.2\textwidth]{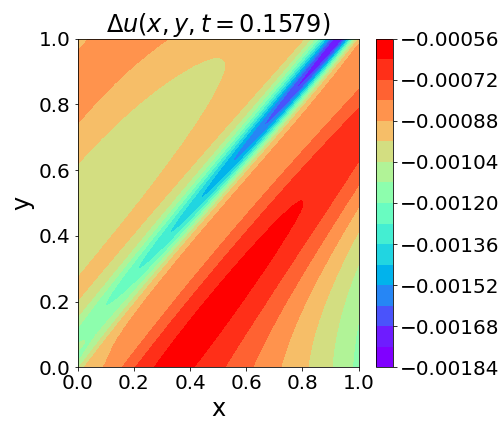}
  \label{fig:burgers2d_t=0.1579noerror_noise}
  }
  \subfloat[]{
  \includegraphics[width=0.2\textwidth]{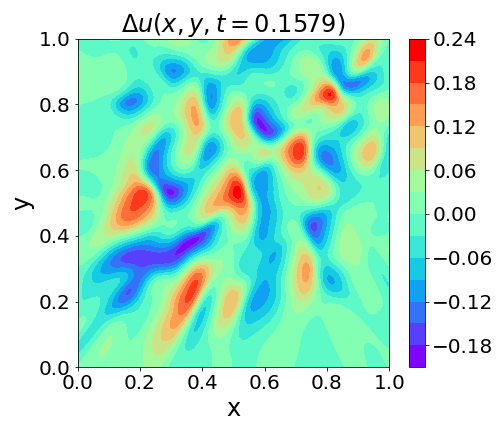}
  \label{fig:burgers2d_t=0.1579nosmoothing_noise}
  }
  \subfloat[]{
  \includegraphics[width=0.2\textwidth]{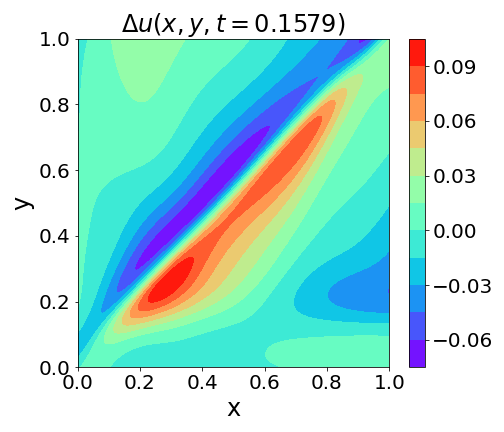}
  \label{fig:burgers2d_t=0.1579gp_noise}
  }
  \subfloat[]{
  \includegraphics[width=0.2\textwidth]{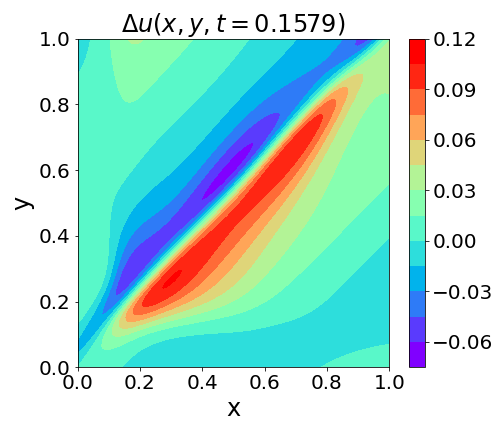}
  \label{fig:burgers2d_t=0.1579sgp_noise}
  } \\
  \subfloat[]{
  \includegraphics[width=0.2\textwidth]{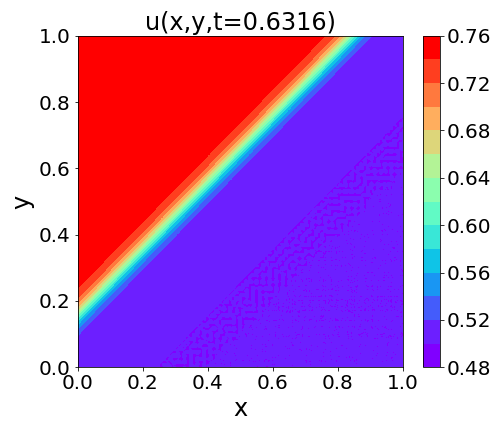}
  \label{fig:burgers2d_t=0.6316exact}
  }
  \subfloat[]{
  \includegraphics[width=0.2\textwidth]{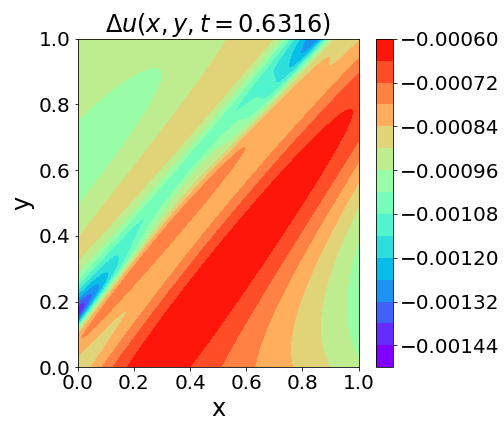}
  \label{fig:burgers2d_t=0.6316noerror_noise}
  }
  \subfloat[]{
  \includegraphics[width=0.2\textwidth]{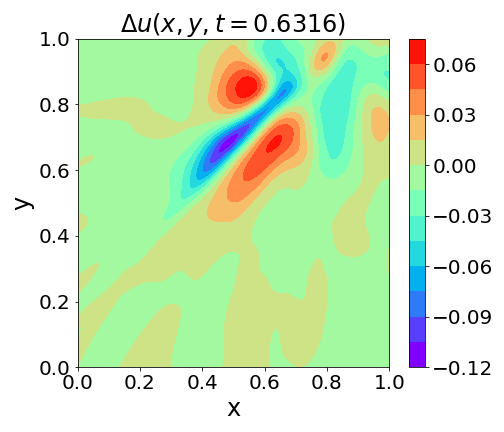}
  \label{fig:burgers2d_t=0.6316nosmoothing_noise}
  }
  \subfloat[]{
  \includegraphics[width=0.2\textwidth]{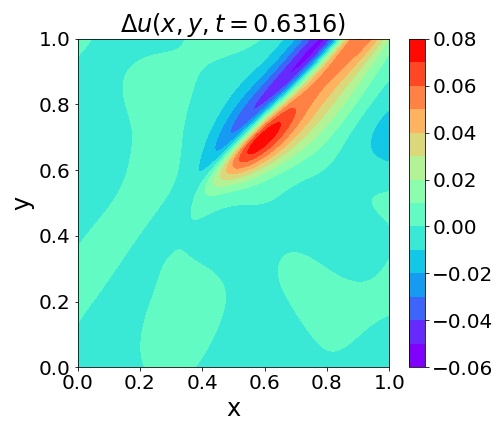}
  \label{fig:burgers2d_t=0.6316gp_noise}
  }
  \subfloat[]{
  \includegraphics[width=0.2\textwidth]{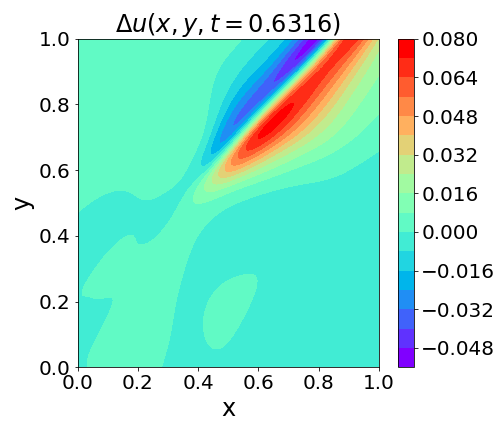}
  \label{fig:burgers2d_t=0.6316sgp_noise}
  } \\
    \subfloat[]{
  \includegraphics[width=0.2\textwidth]{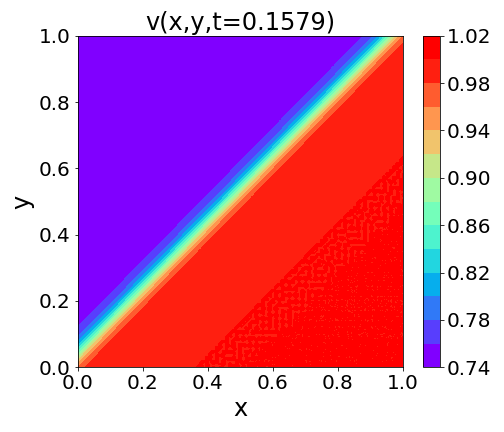}
  \label{fig:burgers2d_t=0.1579exact}
  }
  \subfloat[]{
  \includegraphics[width=0.2\textwidth]{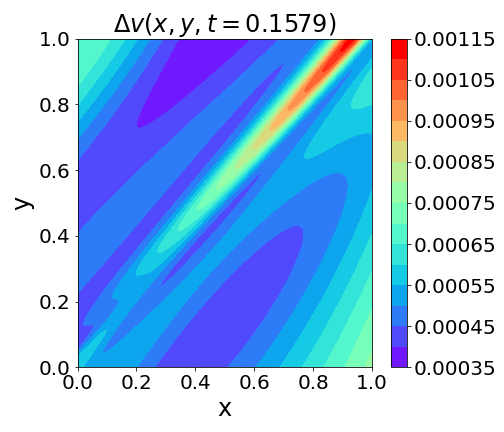}
  \label{fig:burgers2d_t=0.1579noerror_noise}
  }
  \subfloat[]{
  \includegraphics[width=0.2\textwidth]{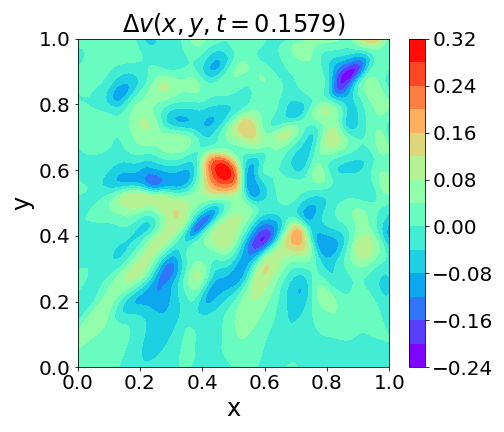}
  \label{fig:burgers2d_t=0.1579nosmoothing_noise}
  }
  \subfloat[]{
  \includegraphics[width=0.2\textwidth]{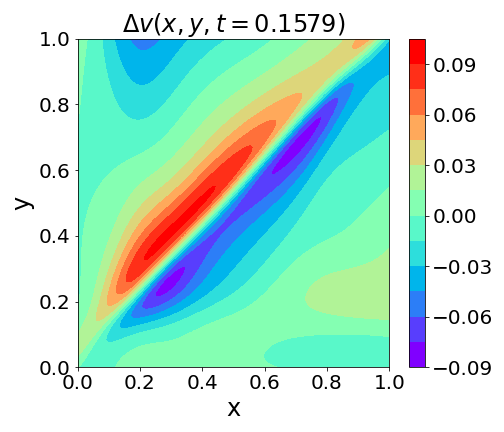}
  \label{fig:burgers2d_t=0.1579gp_noise}
  }
  \subfloat[]{
  \includegraphics[width=0.2\textwidth]{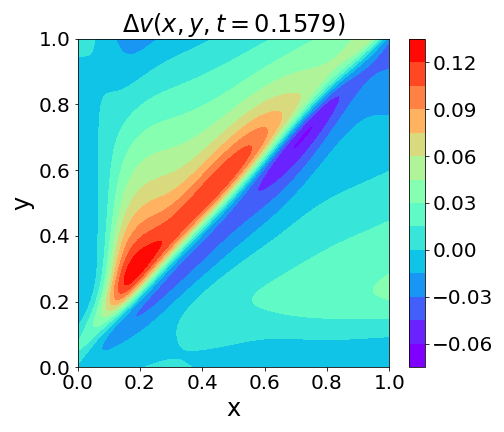}
  \label{fig:burgers2d_t=0.1579sgp_noise}
  } \\
  \subfloat[]{
  \includegraphics[width=0.2\textwidth]{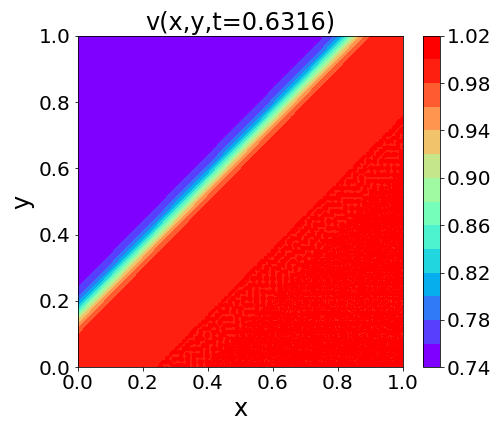}
  \label{fig:burgers2d_t=0.6316exact}
  }
  \subfloat[]{
  \includegraphics[width=0.2\textwidth]{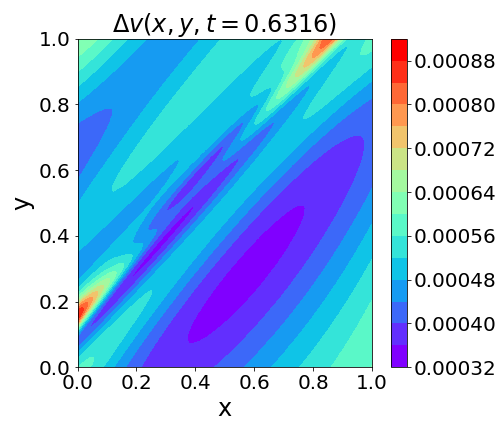}
  \label{fig:burgers2d_t=0.6316noerror_noise}
  }
  \subfloat[]{
  \includegraphics[width=0.2\textwidth]{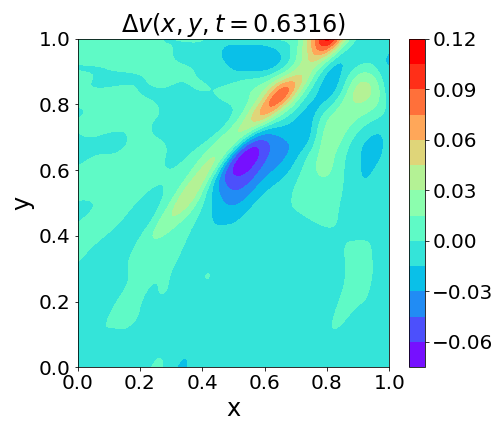}
  \label{fig:burgers2d_t=0.6316nosmoothing_noise}
  }
  \subfloat[]{
  \includegraphics[width=0.2\textwidth]{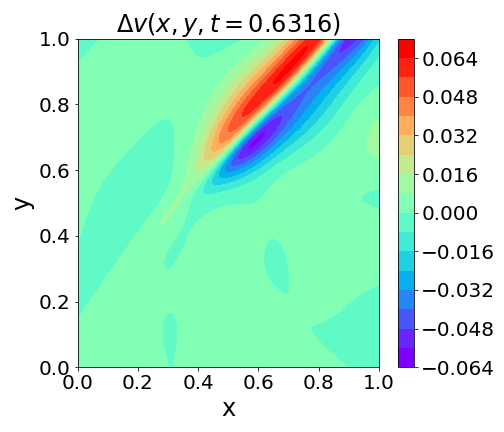}
  \label{fig:burgers2d_t=0.6316gp_noise}
  }
  \subfloat[]{
  \includegraphics[width=0.2\textwidth]{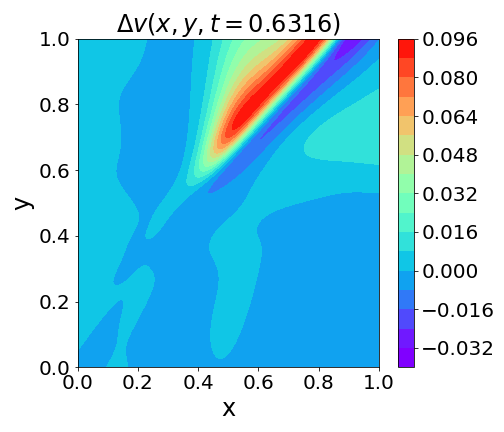}
  \label{fig:burgers2d_t=0.6316sgp_noise}
  }
  \caption[]{The leftmost column shows the exact solution for 2D Burgers' equation at $t = 0.1579$ (first and third row from top for $u(x,y,t)$) and $t= 0.6316$ (second and fourth row from top for $v(x,y,t)$). The remaining columns show the point-wise error in PINN-evaluated solution for noise-free PINN (second column), noisy PINN without smoothing (third column), GP-PINN (fourth column), and SGP-PINN (final i.e. fifth column). Except the vanilla PINN (no error), all models were trained with data sampled from the initial timeslice corrupted with additive Gaussian noise with zero mean and $\sigma = 0.5$. }
  \label{fig:burgers2d_timeslices}
\end{figure}

One of the noticeable aspects of the 2D Burgers' equations is the presence of a shockwave front at $4y - 4x = t$ around where both fields experience rather sharp, yet continuous gradients. We show the exact initial condition used to train the noise-free PINN in Figures~\ref{fig:burgers2d_u_ICexact} and \ref{fig:burgers2d_v_ICexact}. When corrupted with noise and left unsmoothened, the shockwave feature is almost completely lost and as can be seen from Figure~\ref{fig:burgers2d_timeslices} (third column from the left), the network struggles to retrieve the shockwave front for latter timeslices as well. On the other hand, while GP and SGP to some extent recovers the initial field distributions, the gradients near the shockwave front are oversmoothed. It should be noted that this oversmoothing is not due to some limitation of GP or SGP itself, but rather the choice of samples used to train these processes. Being agnostic to the physical distribution, the enforcing points on the initial timeslice are uniformly sampled, which led to an under-representation of sharp gradients near the shockwave front. As a consequence to the missing perception of sharpness around the shockwave front, the GP-PINN and SGP-PINN accumulate their largest deviations in the latter timeslices around the shockwave front, as seen in Figure~\ref{fig:burgers2d_timeslices} (fourth and fifth columns from the left). The opposite signs of the errors on the two sides of the shockwave front represent that the PINN-reconstructed solution after GP or SGP smoothing has a more smoothly shifting wavefront.

%% file: tex-paper/6Conclusion.tex
\section{Conclusion}
As it often happens, measurements associated physical processes are subject to errors. When these measurements are used to learn the evolution of a system respecting some underlying physics dictated by a PDE using NNs, these errors can significantly distort the predicted behavior via nonlinear propagation of errors.
In this paper, we explored the behavior of a PINN when it is trained with noise-corrupted datasets. 
Our work shows that deep PDE-solvers can be subject to overfitting and dynamically propagating errors observed on the domain boundaries even when physics-inspired regularizers are introduced to constrain the solution. 
To circumvent this issue, we proposed GP-smoothed deep network that can help recover the system's behavior over a finite space-time domain while providing a controlled prediction and bounded uncertainty. We further showed that the computational complexity of fitting a Gaussian Process can be significantly reduced by incorporating sparsely choosing inducing points for sparse GPs.
This opens up opportunities to explore uncertainty propagation in predictive estimation using cPINNs or cPINN-like architectures as well as learning an optimal policy of selecting sparsely chosen inducing points.

\section*{Acknowledgement} 
The research for C.B. and L.M was supported in part  by NIH-R01GM117594, by the Peter O'Donnell Foundation,  and in part from a grant from the Army Research Office accomplished under Cooperative Agreement Number W911NF-19-2-0333. The work of L.M. was additionally supported by the the Moncreif Summer Undergraduate Internship Program. The work of T.A. and A.R. is supported by U.S. Department of Energy, Office of High Energy Physics under Grant No. DE-SC0007890. Studies performed by A.R. utilize resources supported by the National Science Foundation’s Major Research Instrumentation program, grant 1725729, as well as the University of Illinois at Urbana-Champaign